%% file: main.tex
\pdfoutput=1
\documentclass[11pt]{article}

\usepackage[preprint]{acl}

\usepackage{times}
\usepackage{latexsym}

\usepackage[T1]{fontenc}
\usepackage[utf8]{inputenc}

\usepackage{microtype}

\usepackage{inconsolata}

\usepackage{graphicx}

\usepackage{amsmath}

\usepackage{xcolor}
\definecolor{MutedCoral}{RGB}{220, 110, 100}       % Softer coral for readability
\definecolor{SeafoamGreen}{RGB}{110, 180, 150}     % Muted seafoam green
\definecolor{DustyBlue}{RGB}{80, 130, 180}         % Darker dusty blue
\definecolor{GoldenSand}{RGB}{230, 185, 110}       % Slightly subdued golden sand
\definecolor{BlushPink}{RGB}{230, 160, 180}        % Less bright blush pink
\definecolor{MistyLavender}{RGB}{160, 130, 180}    % Darker misty lavender
\definecolor{TealBreeze}{RGB}{90, 170, 200}        % Less bright teal breeze
\definecolor{OliveDrab}{RGB}{150, 120, 70}         % Warmer, earthier green-brown
\definecolor{PalePeach}{RGB}{225, 190, 170}        % Darkened pale peach for visibility

\usepackage{multirow}

\usepackage{listings}

\usepackage{etoolbox} % For toggling on/off
\usepackage{tikz} % For drawing the circle

\title{Towards Conditioning Clinical Text Generation for User Control}

\author{Osman Alperen Kora\c{s}$^{1}$  \hspace{0.05em} Rabi Bahnan$^{1}$  \hspace{0.05em}  Jens Kleesiek$^{1,2,3,4}$  \hspace{0.05em}  Amin Dada$^{1}$\\
        $^1$Institute for AI in Medicine (IKIM), University Hospital Essen (AöR), Essen, Germany \\
        $^2$Cancer Research Center Cologne Essen (CCCE), West German Cancer Center Essen \\ 
        University Hospital Essen (AöR), Essen, Germany \\
        $^3$German Cancer Consortium (DKTK, Partner site Essen), Heidelberg, Germany \\
        $^4$Department of Physics, TU Dortmund, Dortmund, Germany}

\begin{document}
\maketitle

\input{src/00_abstract}

\input{src/01_introduction}
\input{src/02_related_work}
\input{src/03_methodology}
\input{src/04_experiments}
\input{src/05_results_and_discussion}
\input{src/06_conclusion}


% TODO: FIX THE DOI OF ALL CITATIONS IN CUSTOM
\bibliography{anthology,custom}

\appendix

\input{src/07_appendix}

\end{document}

%% file: src/00_abstract.tex
\begin{abstract}
Deploying natural language generation systems in clinical settings remains challenging despite advances in Large Language Models (LLMs), which continue to exhibit hallucinations and factual inconsistencies, necessitating human oversight. This paper explores automated dataset augmentation using LLMs as human proxies to condition LLMs for clinician control without increasing cognitive workload. On the BioNLP ACL'24 Discharge Me! Shared Task, we achieve new state-of-the-art results with simpler methods than prior submissions through more efficient training, yielding a 9\% relative improvement without augmented training and up to 34\% with dataset augmentation. Preliminary human evaluation further supports the effectiveness of our approach, highlighting the potential of augmenting clinical text generation for control to enhance relevance, accuracy, and factual consistency.
\end{abstract}

%% file: src/01_introduction.tex
\section{Introduction}
Large language models (LLMs) like OpenAI's GPTs~\citep{openai2024gpt4technicalreport, brown2020languagemodelsfewshotlearners}, Google's PaLM~\citep{anil2023palm2technicalreport} and Gemini~\citep{geminiteam2024geminifamilyhighlycapable}, and lately Meta's Llama~\citep{touvron2023llamaopenefficientfoundation,touvron2023llama2openfoundation,dubey2024llama3herdmodels} have shown remarkable versatility across a wide range of applications, including healthcare~\citep{singhal2023expertlevelmedicalquestionanswering,Huang2024PatientRepresentingPP}. In clinical environments, LLMs offer potential for automating tasks such as summarizing clinical notes, supporting diagnostic decisions, and streamlining patient communication~\citep{ijerph20043378,soleimani2024practical,ruinelli-etal-2024-experiments,Liu2023UtilityOC, Patel2023ChatGPTTF, van2024adapted, Zaretsky2024GenerativeAI,Were2010CreationAE}. However, deploying AI in clinical settings remains a critical challenge due to the high cost of hallucinations, factual inconsistencies, and misinterpretations~\citep{10.1145/3571730, Lin2024TowardsTL,tang2023evaluating, dada2024clue}. Even minor inaccuracies in AI-generated clinical content can lead to severe consequences, such as misdiagnoses, incorrect treatments, or harmful patient outcomes. Ethical considerations further complicate this process, calling for clinicians to hold accountability for medical decisions through rigorous oversight~\citep{Mesk2023TheIF,PMID:38285984}. At the same time, verifying AI-generated content introduces new cognitive burdens, potentially negating the intended efficiency gains of automation.
\begin{figure}[t!]
    \centering
   \includegraphics[width=\columnwidth]{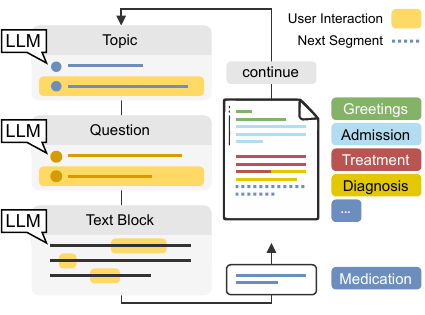}
  \caption{An interactive workflow showcasing topic-level generation control. The LLM is prompted once with the respective context to begin structured generation. After each element, generation is paused, enabling users to sequentially refine content by editing LLM-suggested topic headings, questions, and text blocks. The generation resumes with user-verified content.\vspace{-0.15cm}}
  \label{fig:interactive_generation_workflow}
\end{figure}
As clinicians already face high cognitive workloads, 
addressing this paradox is essential to harness AI's potential in clinical settings without increasing risks or workloads. To strike this balance, AI systems must provide clinicians with control and transparency, ensuring outputs align with clinical contexts, communication styles, and guidelines. This paper explores whether augmenting traditional datasets to condition LLMs for controlled clinical text generation is a viable solution. Specifically, we introduce a system that separates stylistic and content-related requirements, breaking down generation into distinct, manageable writing subtasks. This reduces the complexity of content creation and human verification through a separation of concerns, empowering users to impose authoring guidelines and dynamically guide the process while moving away from black-box models that limit clinician involvement. 

Since traditional datasets do not inherently support such user control, we augment them with authoring guidelines and topic segmentation to condition models for style and content control. Automated evaluation suggests that our approach significantly enhances relevance, accuracy, and factual consistency, highlighting the potential of such augmentations for clinical text generation. Furthermore, we find that traditional instruction-tuning for clinical text generation can be significantly improved through optimized hyperparameter settings, without increasing the compute budget.  Our key contributions\footnote{All source code will be released upon paper acceptance.} are: 

\textbf{New state-of-the-art.} We set a new state-of-the-art on the BioNLP ACL'24 Shared Task 'Discharge Me!' challenge through efficient training, while being simpler and requiring less training compute. 

\textbf{Dataset Augmentation.} We propose methods using LLMs as human proxies to augment traditional datasets, enabling granular control over content and style in clinical text generation. This yields a 34\% relative improvement over prior state-of-the-art, representing a lower bound on potential gains.

\textbf{Human Evaluation.} We conduct preliminary human evaluation, validating the effectiveness and automated evaluation of our approach.

%% file: src/02_related_work.tex
\section{Related Work}
In recent years, there has been increasing research into LLM-based clinical text generation, highlighting its potential in generating discharge summaries \citep{ando2022artificial, ellershaw2024automated, clough2024transforming, dubinski2024leveraging}, brief hospital courses~\citep{hartman2022day, hartman2023method, searle2023discharge}, and radiology reports \citep{alfarghaly2021automated, wang2023r2gengpt, yang2023radiology}. One study even shows that physicians often prefer AI-generated clinical texts over manually written ones~\citep{van2024adapted}. However, most of these approaches have treated clinical text generation as an end-to-end generation task, without offering user intervention and control. A recent example is the BioNLP ACL'24 Shared Task 'Discharge Me!'~\citep{xu-etal-2024-overview}, focusing on generating discharge summaries. However, the complexity of clinical texts, which often require external sources of information and are subject to individual guidelines and writing styles, makes end-to-end generation less feasible in practice. These challenges point to the need for more flexible generation allowing users to control specific aspects of the output, such as content and style.

Controlled text generation (CTG) is a growing area of research aimed at providing users with more influence over the generated content. This involves integrating specific control conditions, such as enforcing a professional tone or ensuring the use of domain-specific terminology, while maintaining fluency, coherence, and relevance in the generated text~\citep{zhang2023survey}. Prior work in this area has explored different control mechanisms that can be directly applied to clinical text generation, such as structure control~\citep{yang-klein-2021-fudge, zou2021controllable}, general style control~\citep{keskar2019ctrl}, and personal style control~\citep{tao2024cat}.

Moreover, recent studies have explored using Question-Answer (QA) pairs as a blueprint to guide the text generation process. This approach has been shown to reduce hallucinations and improve the factual consistency of generated content~\citep{narayan-etal-2023-conditional, huot-etal-2023-text}. It is based on the Question Under Discussion (QUD) theory~\citep{roberts:2012:information}, which states that all utterances in a discourse~\citep{van1995discourse} serve to answer either implicit or explicit questions. Building on these insights, we adapt the QUD framework for clinical document generation by framing clinical documents as responses to implicit questions arising from their intended purpose. These questions are typically addressed in a structured manner, even when the document appears unstructured. Using fine-grained topic segmentation, we aim to uncover this underlying structure by generating headings and QUDs, with corresponding text segments acting as their answers. This approach aligns topics with specific writing subtasks, simplifying the generation process while preserving the structured nature of clinical documentation.

%% file: src/03_methodology.tex
\section{Conditioning Clinical Text Generation for User Control}
We explore two strategies to condition Large Language Models (LLMs) for controlled clinical text generation: (a) topic-level structured generation and (b) authoring guidelines. However, implementing these strategies reveal limitations in traditional datasets, particularly in clinical text generation.

\subsection{Limitations in Traditional Datasets}
\label{sec:limitations-in-traditional-datasets}
Traditionally, training datasets are built on the assumption that more data leads to better generalization. However, in conditional text generation (e.g., summarization) the same task can be completed in multiple stylistically distinct but equally valid ways. Despite this, evaluation benchmarks typically provide only a single reference text, failing to account for the diversity of valid solutions or specifying which variant of task completion is expected. This issue is made by design and cannot be resolved simply by increasing dataset size. Consequently, models are evaluated against a single stylistic realization of a task, potentially skewing evaluation results.

This is particularly evident in clinical datasets, where medical documents exhibit significant differences in quality, format, and style --- even within the same task~\citep{POLLARD201339,edwards2014association,Hultman2019ChallengesAO}. Discharge summaries, for instance, are often compiled from pre-existing records authored by multiple individuals. Contents are often copied across teams, departments, or wards, each adhering to distinct conventions shaped by institutional workflows, time constraints, and resource limitations, leading to inherent stylistic inconsistencies, which is further amplified by situational pressures. Moreover, medical professionals exhibit highly distinctive writing styles, often to the extent that colleagues can recognize one another solely by their writings. Consequently, even within a single discharge summary, different sections may reflect different writing styles, making it impossible to reliably infer the appropriate writing style for one section from the remaining document.

This issue has been largely overlooked in prior research, and to our knowledge, no systematic study has investigated its implications. In particular, the extent to which evaluation metrics are sensitive to stylistic variations, and the degree to which stylistic features emerge due to spurious correlations in input data, remains unclear.
To ensure models can be held accountable for stylistic deviations, we extend the task definition by integrating authoring guidelines into the input context, conditioning the model to adhere to explicit stylistic requirements. This introduces a clear separation of concerns: synthesizing clinically relevant information to complete the task (\textbf{content}) and ensuring conformity to specified conventions (\textbf{style}). Moreover, explicitly conditioning models on authoring guidelines facilitates the emergence of stylistic features through user control, rather than spurious correlations, enabling clinicians to specify institutional or personalized guidelines during inference and promising better generalization.

Another limitation with traditional datasets is their end-to-end design, where the entire output is generated in a single step from the input. This inherently restricts user intervention and control during generation. To train models for (a) controllable and (b) intervenable generation (cf. Fig.~\ref{fig:interactive_generation_workflow}), we need models to sequentially generate output block by block in a structured format with (a) guidance signals to steer the generation of individual blocks and (b) control sequences to start and terminate individual blocks. To address this, we explore fine-grained topic segmentation to structure target texts $t_i$ into XML-formatted sequences.

\subsection{Topic-Level Generation Control}
\label{sec:topic-level-generation-control}
To train models for controllable and intervenable generation,
we tasked Llama 3.1 70B Instruct with fine-grained topic segmentation of target texts $t_i$. The LLM is prompted to segment texts $t_i = (t_i^1, ..., t_i^n)$ into smaller text blocks $t_i^k$, while generating topic-specific headings $\mathring{h}_i^k$ and questions $\mathring{q}_i^k$ for each segment. The output is requested as an XML-structured sequence $$\mathring{seg}(t_i) = \left[
\mathring{h}_i^1, \mathring{q}_i^1, \mathring{t}_i^1, ..., \mathring{h}_i^n, \mathring{q}_i^n, \mathring{t}_i^n\right],$$ in the following format:
\begin{quote}
\texttt{<topic>$\mathring{h}_i^1$</topic>\newline
<question>$\mathring{q}_i^1$</question>\newline
<span>$\mathring{t}_i^1$</span>\newline
$\dots$\newline
<topic>$\mathring{h}_i^n$</topic>\newline
<question>$\mathring{q}_i^n$</question>\newline
<span>$\mathring{t}_i^n$</span>}
\end{quote}

While the headings and questions serve as guidance signals during generation, the XML tags serve as control sequences to stop, adjust and continue generation in each distinct phase  (Fig.~\ref{fig:interactive_generation_workflow}). The prompt (Tab.~\ref{table:topic_segmentation_annotation_example}) is designed to enforce fine-grained topic segmentation, without imposing a particular concept of topics or questions. It's summarized as follows: (1) a new segment should begin when the clinical focus changes, which we associate with a new writing subtask, (2) headings $\mathring{h}_i^k$ should summarize their respective segment, which we equate with the topic, and which (3) should be rephrased as a question $\mathring{q}_i^k$ answered by the respective segment $t_i^k$, which we consider to be the Question Under Discussion (QUD) of said segment. The remaining guidelines are provided to ensure standardization.

A post-processing step then restores the original character sequences $t_i^1, ..., t_i^n$ from $t_i$ for each generated text block $\mathring{t}_i^1, \dots, \mathring{t}_i^n$ (see Appendix~\ref{sec:topic-segmentation-post-processing}), as the LLM generated text blocks $\mathring{t}_i^k$ may not replicate the input $t_i$. The final output is denoted as: $$seg(t_i) = \left[
\mathring{h}_i^1, \mathring{q}_i^1, t_i^1, ..., \mathring{h}_i^n, \mathring{q}_i^n, t_i^n\right].$$

However, to avoid introducing inconsistencies between headings, questions, and text blocks, this step is applied selectively only to those segmentations $\mathring{seg}(t_i)$, which introduce only minor alterations to the input (see Appendix~\ref{sec:topic-segmentation-post-processing}).

\subsection{Authoring Guidelines}
\label{sec:authoring-guidelines}
From a user perspective, authoring guidelines govern the requirements a document must comply with. These may range from stylistic features to structural constraints. Conditioning text generation on such guidelines may therefore not only improve alignment of model outputs with user intent, but also provide greater control over generation. However, traditional datasets often lack such guidelines. In this work, we explore the feasibility of using LLMs to close this gap in clinical datasets.

Specifically, we explore the use of two types of automatically generated authoring guidelines for clinical documents $t_i$, which differ in their formulation: (a) style guidelines, which describe the stylistic features a clinical document should express and (b) writing instructions, guiding a non-specialist in writing a clinical document that serves the intended purpose while expressing the desired stylistic features. To achieve this, Llama 3.1 70B Instruct is prompted independently for each target text $t_i$ as follows:

\textbf{Style Guidelines.} The LLM is prompted to describe the stylistic features of the target text $t_i$, including tone, document format, layout, composition, text structure, use of language (including abbreviations and medical jargon), and intended audience (cf. Tab.~\ref{table:style_guideline_example}).

\textbf{Writing Instructions.} The LLM is prompted to generate markdown-formatted instructions for guiding a non-specialist in replicating the target text $t_i$, including directives on the same stylistic features as above while specifying the purpose, document type and outline (cf. Tab.~\ref{table:writing_instructions_example}).

The LLM prompts are carefully engineered to avoid answer leakage by instructing the LLM to not use terms or phrases from the source text, to not quote or give examples from the patient records, and not to reveal patient-specific details.

\begin{figure}[t!]
  \includegraphics[width=\columnwidth]{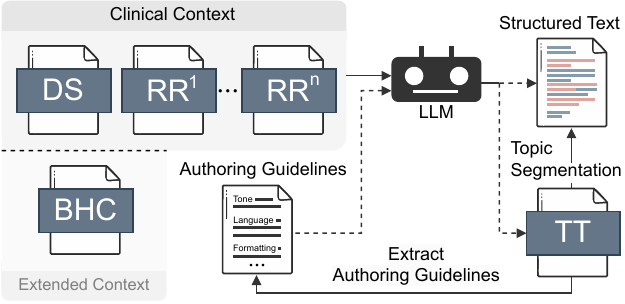}
  \caption{Instruction-tuning pipeline. Dashed lines indicate paths that depend on the training configuration. Models with topic-level control are trained to generate XML-structured text. The extended context is provided only for TT~=~DI. Abbreviations: Discharge Summary~(DS), Radiology Report~(RR), Discharge Instructions~(DI), Brief Hospital Course~(BHC), Target Text~(TT).}
  \label{fig:training-procedure}
\end{figure}

\subsection{Instruction Tuning for Controlled Clinical
Text Generation}
\label{subsec:instruction-tuning-for-controlled-clinical-text-generation}
We utilize the \textit{Discharge Me!} challenge\footnote{\url{https://stanford-aimi.github.io/discharge-me}}, part of the BioNLP ACL'24 Shared Tasks, for training and evaluating our models due to its clinical relevance and challenging nature. Additionally, its leaderboard provides a strong baseline. The task focuses on automating the generation of hospital course summaries and discharge instructions, traditionally time-intensive tasks for clinicians. 

\textbf{Dataset.} The dataset consists of 109,168 discharge summaries from the MIMIC-IV dataset, each containing a Brief Hospital Course (BHC) and a Discharge Instructions (DI) section. It is divided into training (68,785), validation (14,719), phase I test (14,702), and phase II test (10,962) sets. The BHC section is typically found in the middle of the discharge summary, following details on patient history and treatments during the current visit. The DI section is generally located at the end of the note. Additionally, each discharge summary is linked to at least one radiology report and typically one ICD chief complaint, along with multiple ICD codes. The DI and BHC sections are removed from the discharge summary, and serve as target texts $t_i$. The clinical input constitutes of the remaining discharge summary (DS)~and~radiology~reports~(RR).

To address the aforementioned limitations (\ref{sec:limitations-in-traditional-datasets}), we generate topic segmentations (\ref{sec:topic-level-generation-control}), style guidelines and writing instructions (\ref{sec:authoring-guidelines}) for each DI and BHC section $t_i$ separately. We employ Llama 3.1 70B Instruct for these tasks, as LLMs have shown to be an effective  substitute for human annotators~\citep{doi:10.1073/pnas.2305016120,perez-etal-2022-red}.

\begin{figure}[t!]
\centering
\begin{tabular}{p{0.9\columnwidth}}
\hline
User Message \\
\hline
\vspace{-1em}
\begin{lstlisting}
{{discharge summary}}
{{radiology report 1}}
         ...
{{radiology report n}}
{{brief hospital course}}
{{authoring guidelines}}
{{instructions}}
\end{lstlisting} \vspace{-1.5em} \\
\hline
Assistant Message \\
\hline
\vspace{-1em}
\begin{lstlisting}
{{output}}
\end{lstlisting} \vspace{-1.5em} \\
\hline
\end{tabular}

\caption{The generic template for $prompt_i(c,g)$ used for instruction-tuning.}
\label{fig:prompt_template}
\end{figure}

\textbf{Instruction-Tuning Prompts.} 
We fine-tune our models with instruction-tuning on completions only using a generic template (cf. Fig.~\ref{fig:prompt_template}) $$prompt_i(c, g) = (user_i(c,g), assistant_i(c)),$$ where $c \in \{ \texttt{none,  \allowbreak topics} \}$ denotes the possible configurations for structuring the generation output for control and $g \in \{ \texttt{none, \allowbreak  style,  \allowbreak instr} \} $ denotes the possible configurations for using authoring guidelines.

\textbf{User Messages.} $user_i(c,g)$ include the clinical context, consisting of the discharge summary $ds_i$ and radiology reports $r_i^1, \dots, r_i^j$. For generating discharge instructions ($di_i$), we additionally include the brief hospital course report ($bhc_i$). If  $g \in \{ \texttt{style,  \allowbreak instr} \}$, we also include the respective authoring guidelines (cf. Fig.~\ref{fig:training-procedure}) and instruct the model to comply.
If $c = \texttt{topics}$,  the model is instructed to generate XML-structured output for topic-level structured generation.
Separate instructions are provided for the DI and BHC generation tasks.

\textbf{Assistant Messages.} $assistant_i(c)$ contains the desired output, which is the plain target text $t_i \in \{di_i, bhc_i\}$ for $c = \texttt{none}$, or the XML-structured output $seg_i(t_i)$ for $c = \texttt{topics}$ (Sec.~\ref{sec:topic-level-generation-control}).

%% file: src/04_experiments.tex
\section{Experiments and Evaluation}
\vspace{-0.4em}
\subsection{Experimental Setup and Baselines}

We fine-tune \textit{Llama 3 8B Instruct} on the training split of the \textit{Discharge Me!} challenge dataset with instruction tuning using $prompt_i(c, g)$ for all possible configurations (see Section~\ref{subsec:instruction-tuning-for-controlled-clinical-text-generation}). See Appendix~\ref{subsec:Training-Details} for training details. This model is chosen to maintain a fair comparison with~\citet{damm-etal-2024-wispermed}, who placed first on the leaderboard by employing a Dynamic Expert Selection (DES) system that included \textit{Llama 3 8B Instruct} as one of its smaller models. We evaluate on the test-phase-2 split used to determine the final leaderboard rankings. The Top 3 leaderboard entries serve as the state-of-the-art baseline for the Brief Hospital Course (BHC) and Discharge Instructions (DI) generation tasks.

\textbf{BASE} denotes our model which is trained without any data augmentations ($c = \texttt{none}$, $g = \texttt{none}$). It serves as a baseline for our other models. Models trained with authoring guidelines ($g \in \{\texttt{style}, \texttt{instr}\})$ are indicated with \textbf{\textsc{w/STYLE}} or \textbf{\textsc{w/INSTR}} respectively. Similarly, models trained on structured output ($c = \texttt{topics}$) are indicated with \textbf{\textsc{w/TOPICS}}.

In addition, we prompt the stronger base model \textit{Llama 3.3 70B Instruct} with user messages $user_i(c, g)$ zero-shot and three-shot to assess the gains provided by dataset augmentations without any fine-tuning.

\subsection{Automated Evaluation}
\label{sec:automated-evaluation}
All our models are evaluated using the code provided by the \textit{Discharge Me!} challenge\footnote{https://github.com/Stanford-AIMI/discharge-me/scoring}, which employs a comprehensive set of metrics (see Appendix~\ref{sec:automated-metrics-explanation}) to assess lexical similarity (\textbf{BLEU\nobreakdash-4}, \textbf{ROUGE\nobreakdash-1}, \textbf{ROUGE\nobreakdash-2}, \textbf{ROUGE\nobreakdash-L}, \textbf{METEOR}), semantic similarity (\textbf{BERTScore}), factual consistency (\textbf{AlignScore}), and the clinical relevance and correctness (\textbf{MEDCON}) of the generated texts $\hat{t}_i$ in comparison to the gold-standard target texts $t_i$. It was reported, that this ensemble resulted in rankings that aligned well with clinician evaluation \cite{xu-etal-2024-overview}.
For evaluation, we first complete the BHC task and then use the output to generate the DI section. Greedy decoding is used for inference. For models w/\textsc{topics}, which generate XML-structured outputs $seg(\hat{t}_i)$, the output is parsed into plain text by joining the spans $\hat{t}_i^1, \dots, \hat{t}_i^n$ with white spaces to retrieve the final model output.

To simulate user-control, we adopt a methodology (see Appendix~\ref{sec:automated-metrics-explanation}) inspired by prior work~\citep{10.1145/3660810,10606356}, leveraging LLMs as proxies for human evaluators to automate evaluation on existing benchmarks. Specifically, an LLM acts as a proxy for the original authors of the DI and BHC sections by generating authoring guidelines and providing topic guidance.
For simplicity, topic guidance is provided indirectly and non-interactively, without refining outputs to match the target text, establishing a lower-bound baseline for performance.

\subsection{Human Evaluation}
The evaluation of interactive, user-controlled models would ideally involve a user study, where users engage with the models to generate DI and BHC sections. However, conducting such a study at scale is beyond the scope of this exploratory study, as it is too resource-intensive and time-consuming. We therefore complement our automatic evaluation with two preliminary human evaluations to assess the effectiveness of our approach and the quality of the dataset augmentations.

The first evaluation assessed whether our models generate clinically appropriate outputs when provided with human-written guidelines and whether automatic evaluation metrics align with human judgment. An advanced medical student in his final clinical year, serving as a domain expert, dedicated 95 hours and 13 minutes to manually authoring 200 guidelines for the DI and BHC sections of 100 randomly sampled discharge summaries from the test-phase-2 split of the 'DischargeMe!' dataset.  
While no fixed template was imposed, the expert was encouraged to consider elements such as document type, content coverage, structure, formatting, tone, use of language, complexity, and technicality.
To ensure the guidelines captured clinically relevant stylistic and structural directives, while authentically reflecting human-written guidelines, the expert was instructed to: (1)~Write naturally, following personal preferences, rather than adhering to rigid templates.
(2)~Provide guidance enabling a non-medical layman to write the target text solely based on the discharge summary.
(3)~Avoid medical jargon and patient-specific details, while capturing key clinical writing conventions.

The second evaluation assessed the quality of LLM-generated topic segmentations, specifically the topic accuracy, question validity, and text block appropriateness. 500 discharge summaries were sampled from the post-processed subset of the training split of the 'DischargeMe!' dataset for this purpose, yielding a total of 1000 segmentations $seg(t_i)$. For each $t_i$,  one segment $[h_i^j, q_i^j, t_i^j]$ was randomly selected for assessment. The same medical expert then dedicated 26 hours and 27 minutes to evaluating each segment through a two-step process (see Appendix~\ref{sec:human-evaluation-of-topic-segmentations}).

%% file: src/05_results_and_discussion.tex
\begin{table}[t!]
\centering
\resizebox{\columnwidth}{!}{
\begin{tabular}{lccr}
\hline
 & 0-shot & 3-shot & RI \\
\hline
Llama 3.3 70B Instruct & 0.175 & \textbf{0.210} & +20\% \\
\hspace{3mm} w/\textsc{style} & 0.184 & \textbf{0.215} & +17\% \\
\hspace{3mm} w/\textsc{instr} & 0.210 & \textbf{0.223} & +6\% \\
\hspace{3mm} w/\textsc{topics} & 0.226 & \textbf{0.227} & +0\% \\
\hspace{3mm} w/\textsc{style} w/\textsc{topics} & \textbf{0.225} & \textbf{0.225} & +0\% \\
\hspace{3mm} w/\textsc{instr} w/\textsc{topics} & \textbf{0.230} & \textbf{0.230} & +0\% \\
\hline
\end{tabular}
}
\caption{Overall evaluation results of Llama 3.3 70B Instruct with zero-shot and three-shot prompting. Relative Improvements (RI) are rounded to integers. See Tab.~\ref{table:task_performance_llama} for detailed results. \vspace{-0.2cm}}
\label{table:llama_results}
\end{table}

\begin{table*}[ht!]
\centering
\resizebox{\textwidth}{!}{
\begin{tabular}{lccccccccc}
\hline
 & Overall & BLEU & R-1 & R-2 & R-L & BS & METEOR & AS & MEDCON \\
\hline
WisPerMed& \textbf{0.332} &\textbf{ 0.124} & \textbf{0.453} & \textbf{0.201} & \textbf{0.308} & \textbf{0.438 }& \textbf{0.403} & \textbf{0.315} & \textbf{0.411} \\
HarmonAiLab@Yale & 0.300 & 0.106 & 0.423 & 0.180 & 0.284 & 0.412 & 0.381 & 0.265 & 0.353 \\
aehrc & 0.297 & 0.097 & 0.414 & 0.192 & 0.284 & 0.383 & 0.398 & 0.274 & 0.332 \\
\hline
BASE & 0.363 & 0.168 & 0.483 & 0.255 & 0.345 & 0.472 & 0.362 & 0.359 & 0.460 \\
\hspace{3mm} w/\textsc{style} & 0.399 & 0.202 & 0.526 & 0.289 & 0.382 & 0.508 & 0.404 & 0.383 & 0.495 \\

\hspace{3mm} w/\textsc{instr} & 0.420 & 0.224 & 0.547 & 0.310 & 0.404 & 0.527 & 0.428 & 0.403 & 0.515 \\

\hspace{3mm} w/\textsc{topics} & 0.403 & 0.195 & 0.524 & 0.287 & 0.384 & 0.503 & 0.414 & 0.402 & 0.517 \\

\hspace{3mm} w/\textsc{style} w/\textsc{topics} & 0.436 & 0.226 & 0.562 & 0.319 & 0.421 & 0.539 & 0.444 & \underline{\textbf{0.429}} & 0.548 \\

\hspace{3mm} w/\textsc{instr} w/\textsc{topics} & \underline{\textbf{0.445}} & \underline{\textbf{0.238}} & \underline{\textbf{0.571}} & \underline{\textbf{0.327}} & \underline{\textbf{0.429}} & \underline{\textbf{0.548}} & \underline{\textbf{0.463}} & 0.426 & \underline{\textbf{0.556}} \\

\hline
\end{tabular}
}
\caption{Evaluation results of the \textit{Discharge Me!} leaderboard leaders WisPerMed~\citep{damm-etal-2024-wispermed}, HarmonAiLab@Yale~\citep{socrates-etal-2024-yale} and aehrc~\citep{liu-etal-2024-e}, and our instruction-tuned models on the test set (phase 2). Bold indicates best scores in each block. In addition, underscoring indicates the overall best score. Figure~\ref{fig:relative_improvments_heatmap} shows relative improvements. Table~\ref{table:task_performance} breaks down performance by task. Abbreviations: BERTScore (BS), AlignScore (AS).}
\label{table:main_results}
\end{table*}

\section{Results and Discussion}

In this section, we analyze the impact of our data augmentation strategies on general instruction-tuned LLMs, evaluate the efficiency of our state-of-the-art training approach, and assess how conditioning text generation for user control enhances clinical text generation. We further present human evaluation results, validating the effectiveness of our approach.

\subsection{Impact of Data Augmentations on General Instruction-Tuned LLMs} 
\label{subsec:user-control-in-general-instruction-tuned-llms}
Llama 3.3 70B Instruct performs significantly worse than previous submissions on the DischargeMe! leaderboard (cf. Tab.~\ref{table:llama_results} vs. Tab.~\ref{table:main_results}). Nonetheless, augmenting the input with authoring guidelines and topic guidance yields a $31\%$ relative improvement in the best configuration ($c = \texttt{topics}, g = \texttt{instr})$ over using no data augmentations.
Notably, zero-shot Llama 3.3 70B Instruct w/\textsc{instr} performs on par with the three-shot setting without any dataset augmentations. This suggest that in-context learning effects can be replicated using explicit authoring guidelines with a lot less context tokens (cf. Tab.~\ref{table:dataset_statistics}). Further supporting this, three-shot prompting provides no additional gains over zero-shot prompting when topic guidance is provided (w/\textsc{topics}).

\textbf{Conclusion:} Overall, zero-shot Llama 3.3 70B Instruct achieves only about half of the performance of our models based on Llama 3 8B Instruct (cf. Tab.\ref{table:main_results}), underscoring the importance of augmenting datasets for user control during training.

\subsection{State-of-the-Art Performance with Efficient Training}
Our instruction-tuned baseline model (BASE), trained without dataset augmentations, achieves a new state-of-the-art on the BioNLP ACL'24 \textit{DischargeMe!} leaderboard, outperforming prior submissions across all metrics except METEOR (Tab.~\ref{table:main_results}). BASE achieves a score of $0.363$, surpassing WisPerMed ($0.332$), HarmonAiLab@Yale ($0.300$), and aehrc ($0.297$) --- a relative improvement of $9\%$ over the previous best model. Compared to them, BASE is more efficient:

\textbf{Smaller Trainable Parameter Size.} BASE has only 169M trainable parameters, which is 5-6× fewer than WisPerMed (1046M), Yale (812M), and aehrc (894M).

\textbf{Simpler Methodology.} We instruction-tune Llama 3 8B Instruct, while WisPerMed employs an ensemble of instruction-tuned Llama 3 8B \& 70B Instruct, OpenBioLLM 70B, Mistral 7B Instruct (v0.2), and Phi 3 Mini 128K Instruct. Yale uses an extended training dataset, while aehrc optimizes the clinical input context for downstream tasks. In addition, other submissions use nucleus sampling or 4-beam search, while we decode greedly.

\textbf{Lower Computational Cost.} Considering all individual training setups, our training requires only $56\%$ of Yale’s compute budget, $23\%$ of aehrc’s, and $32\%$ of WisPerMed’s.

Notably, WisPerMed and aehrc also instruction-tuned Llama 3 8B Instruct using similar approaches, yet reported significantly lower scores ($0.253$ and $0.235$, respectively). Our model achieves relative improvements of $43\%$ over WisPerMed's and $54\%$ over aehrc's fine-tuning attempts. A detailed comparative analysis (Appendix~\ref{sec:comparative-analysis-with-existing-approaches}) suggests that our superior performance stems from more efficient training, which includes higher learning rates, rank-stabilized LoRA and SVD-based PISSA.

\textbf{Conclusion:} Our findings demonstrate that more efficient training strategies can yield substantial improvements, even with fewer parameters and lower computational costs, achieving a new state-of-the-art for clinical text generation.

\subsection{Conditioning Text Generation for User Control}
\label{sec:conditioning-text-generation-for-user-control}
\textbf{Authoring Guidelines.} Augmenting datasets with authoring guidelines significantly improves model performance (cf. Table~\ref{table:main_results}, Fig.~\ref{fig:relative_improvments_heatmap}). BASE w/\textsc{style} ($0.399$, $+10\%$) and BASE w/\textsc{instr} ($0.420$, $+16\%$) outperform BASE ($0.363$), demonstrating the potential of augmenting datasets with explicit guidelines. 

Style guidelines enhance lexical similarity (BLEU, ROUGE, METEOR) with $9$–$21\%$ relative improvements, compensating for BASE’s METEOR deficit. Surprisingly, even semantic and fact-based metrics (BERTScore, AlignScore, MEDCON) improve by $7$–$8\%,$ suggesting either (i) these metrics are sensitive to stylistic variances or (ii) automatically generated style guidelines contain spurious features that reinforce factual  and clinical alignment --- an area requiring further research.

Writing instructions consistently outperform style guidelines and match BASE w/\textsc{topics} on fact-based metrics (AlignScore, MEDCON: $+12\%$), despite the latter being conditioned for and provided with topic-level guidance.

\textbf{Topic Guidance.} Providing LLMs conditioned for topic-level control with topic guidance yields overall improvements ($+11\%$) similar to style guidelines, but with fact-based metrics (AlignScore, MEDCON: $+12\%$) contributing more. 
Although BASE w/\textsc{style} w/\textsc{topics} ($+20\%$) performs slightly worse, integrating both authoring guidelines and topic guidance yields further performance gains across all metrics, showing that these strategies are complementary, and evidencing the need for both style- and  content-aware conditioning. Notably, our best model BASE w/\textsc{instr} w/\textsc{topics} ($+22\%$) excels in DI generation, achieving high ROUGE-1 ($0.612$), BERTScore ($0.587$), and MEDCON ($0.594$) scores. (cf. Table~\ref{table:task_performance}).

\textbf{Conclusion:} Overall, we observe that stylistic and content-related guidance is complementary, and that all metrics, even fact-based ones, appear sensitive to stylistic deviations to different degrees. Furthermore, clear instructions, expressing the purpose of stylistic features and the document clearly outperform simple stylistic descriptions.

\subsection{Human Evaluation Results}
We evaluate BASE w/\textsc{instr} on a sample of 100 discharge summaries using human-written authoring guidelines (cf. Tab.~\ref{table:human_evaluation_results}). For cross-validation, we also assess BASE and BASE w/\textsc{instr} using automated evaluation (Sec.~\ref{sec:automated-evaluation}). Results (cf. Tab.~\ref{table:human_evaluation_results}) indicate that the sampled dataset is not significantly easier than the original test set (BASE: $0.365,$ BASE w/\textsc{instr}: $0.415$) . When prompted with human-written guidelines, BASE w/\textsc{instr} ($0.403$) retained $97\%$ of its performance, while maintaining clinical accuracy \& relevance (MEDCON: $-0.8\%$) and improving factual consistency (AlignScore: $+4\%$), underscoring the promise of adopting LLMs for expert annotations in clinical practice.

Evaluating the topic segmentations (cf. Appendix~\ref{sec:human-evaluation-of-topic-segmentations}) reveals that $91.9\%$ of LLM-generated headings ($\mathring{h}_i^j$) correctly match their corresponding text blocks ($t_i^j$). Similarly, $88.4\%$  of the generated questions ($\mathring{q}_i^j$) are appropriately answered by the text block and effectively inquire it's content. The selected text range ($t_i^j$) is deemed accurate in $75.2\%$ of cases, meaning it accurately aligns with the optimal segment boundaries for the suggested heading $\mathring{h}_i^j$ and question $\mathring{q}_i^j$. These results suggest that expert-level accuracy may already be within reach with stronger models or a secondary validation pass to refine the segmentation, in particular segment boundaries. 

\textbf{Conclusion:} Our findings reinforce the idea that LLMs can effectively act as human proxies, even for complex multi-step tasks like topic segmentation,  bringing automation closer to expert-level performance.

%% file: src/06_conclusion.tex
\section{Conclusion and Future Work}
In this work, we explored strategies for conditioning LLMs to give clinicians control over both content and style in clinical text generation. Using the BioNLP ACL'24 Shared Task Discharge Me! as a case study, we demonstrated that augmenting datasets with authoring guidelines and topic segmentation significantly improves accuracy, relevance, and factual consistency. Notably, our findings raise concerns about metrics exhibiting significant sensitivity to stylistic deviations, even when fact-based, warranting further research.

Our preliminary human evaluation suggests that LLMs can serve as proxies for expert annotations, enabling dataset augmentation at scale. By introducing a separation of content and style, we extended the traditional clinical text generation paradigm to facilitate the integration of clinical communication and authoring guidelines. Since such guidelines are crafted once per task, they offer a low-cost enhancement to clinical text generation without adding cognitive burden.

We also establish a new state-of-the-art for conventional clinical text generation on Discharge Me!, surpassing prior submissions while using fewer parameters and significantly lower computational costs. To support further research and real-world adoption, we disclose our methods, allowing hospitals and clinical institutions to adapt these augmentations to their own data and workflows.

While preliminary human evaluation validates the effectiveness of our approach, a systematic study is needed to identify which specific components of authoring guidelines contribute most to downstream performance. Future work should also focus on scaling human evaluation, assessing generalization across diverse clinical datasets, and refining LLM conditioning techniques to improve adaptability to real-world medical documentation workflows. Additionally, user studies should evaluate interactivity and its impact on clinician oversight.

\section{Limitations}
While our approach demonstrates strong performance in clinical text generation, several limitations remain. Our findings rely primarily on automated metrics, with only preliminary human evaluation, making a larger-scale, clinician-in-the-loop assessment essential to validate practical usability and real-world adoption. Additionally, this study does not yet evaluate how interactive clinician involvement impacts cognitive workload and oversight burden. Future work should investigate whether LLM-conditioned generation can reduce verification effort and how user feedback can further refine dataset augmentation to better align with clinical workflows. Lastly, as with all large-scale pre-trained models, our approach inherits biases from its training data, potentially affecting fairness and reliability in clinical decision-making. No work was done to mitigate such bias and assess the clinical implications of these biases to ensure responsible AI deployment in healthcare, and the effects remain unknown.

%% file: src/07_appendix.tex
\begin{figure*}[t!]
   \centering
   \includegraphics[width=1.5\columnwidth]{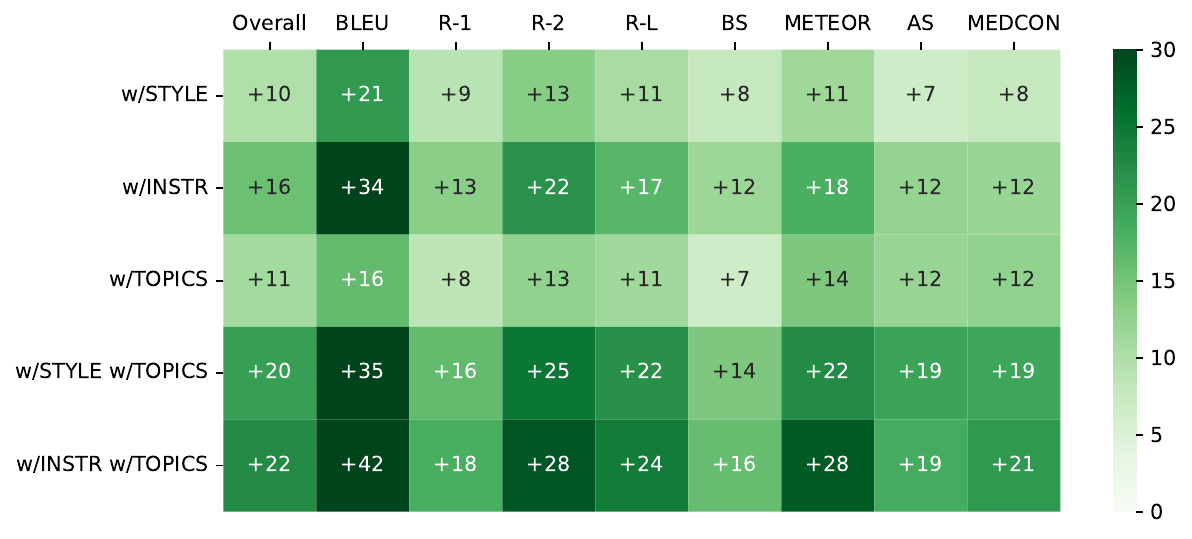}
  \caption{Relative improvement of augmented models against the traditionally instruction-tuned BASE model (cf. Tab.~\ref{table:main_results}).}
  \label{fig:relative_improvments_heatmap}
\end{figure*}

\begin{table*}[t!]
\centering
\resizebox{\textwidth}{!}{
\begin{tabular}{llccccccccc}
\hline
  &  & Overall & BLEU & R-1 & R-2 & R-L & BS & Meteor & AS & MEDCON \\
\hline
\multirow{6}{*}{\rotatebox{90}{\textsc{BHC}}} & BASE  & 0.333 & 0.142 & 0.465 & 0.228 & 0.313 & 0.460 & 0.335 & 0.290 & 0.435 \\
 & \hspace{3mm} w/\textsc{style} & 0.350 & 0.158 & 0.488 & 0.242 & 0.330 & 0.477 & 0.356 & 0.297 & 0.452 \\
 & \hspace{3mm} w/\textsc{instr}  & 0.364 & 0.171 & 0.505 & 0.255 & 0.343 & 0.489 & 0.376 & 0.304 & 0.470 \\
 & \hspace{3mm} w/\textsc{topics}  & 0.356 & 0.149 & 0.487 & 0.239 & 0.333 & 0.471 & 0.369 & 0.317 & 0.482 \\
 & \hspace{3mm} w/\textsc{style} w/\textsc{topics}& 0.385 & 0.178 & 0.524 & 0.268 & 0.367 & 0.504 & 0.396 & \textbf{0.332} & 0.513 \\
& \hspace{3mm} w/\textsc{instr} w/\textsc{topics} & \textbf{0.390} & \textbf{0.183} & \textbf{0.530} & \textbf{0.271} & \textbf{0.370} & \textbf{0.509} & \textbf{0.410} & 0.327 & \textbf{0.517} \\
\hline
\multirow{6}{*}{\rotatebox{90}{\textsc{DI}}} & BASE  & 0.393 & 0.193 & 0.502 & 0.283 & 0.377 & 0.484 & 0.390 & 0.428 & 0.484 \\
 & \hspace{3mm} w/\textsc{style} & 0.448 & 0.247 & 0.565 & 0.337 & 0.434 & 0.539 & 0.452 & 0.469 & 0.538 \\
 & \hspace{3mm} w/\textsc{instr} & 0.476 & 0.277 & 0.589 & 0.366 & 0.464 & 0.565 & 0.480 & 0.503 & 0.560 \\
 & \hspace{3mm} w/\textsc{topics}  & 0.451 & 0.240 & 0.561 & 0.336 & 0.436 & 0.535 & 0.459 & 0.487 & 0.552 \\
 & \hspace{3mm} w/\textsc{style} w/\textsc{topics} & 0.487 & 0.273 & 0.600 & 0.370 & 0.474 & 0.574 & 0.491 & \textbf{0.526} & 0.584 \\
 & \hspace{3mm} w/\textsc{instr} w/\textsc{topics} & \textbf{0.500} & \textbf{0.292} & \textbf{0.612} & \textbf{0.383} & \textbf{0.488} & \textbf{0.587} & \textbf{0.517} & 0.525 & \textbf{0.594} \\
\hline
\end{tabular}
}
\caption{The average scores per metric of our evaluation (Sec.~\ref{sec:conditioning-text-generation-for-user-control}), broken down by the two tasks: discharge instructions (DI) generation and brief hospital course (BHC) generation.}
\label{table:task_performance}
\end{table*}

\begin{table*}[t!]
\centering
\begin{tabular}{lccccccccc}
\hline
 & Overall & BLEU & R-1 & R-2 & R-L & BS & METEOR & AS & MEDCON \\
\hline
BASE & 0.365 & 0.169 & 0.484 & 0.252 & 0.345 & 0.478 & 0.360 & 0.364 & 0.467 \\

\hspace{3mm} w/\textsc{instr} (A) & \textbf{0.415} & \textbf{0.215} & \textbf{0.546} & \textbf{0.303} & \textbf{0.397} & \textbf{0.525} & \textbf{0.419} & 0.400 & \textbf{0.513}  \\
\hspace{3mm} w/\textsc{instr} (H) & 0.403 & 0.193 & 0.524 & 0.288 & 0.390 & 0.514 & 0.392 & \textbf{0.415} & 0.509 \\
\hline
\end{tabular}
\caption{The results of BASE and BASE w/\textsc{instr} evaluated on 100 discharge summaries randomly sampled from the \textit{Discharge Me!} test phase 2 split, once with augmented authoring guidelines (A) and once with human-written authoring guidelines (H).}
\label{table:human_evaluation_results}
\end{table*}

\section{Training Details}
\label{subsec:Training-Details}

We fine-tune \textit{Llama~3~8B~Instruct} on 8~H100 GPUs for 3,000 steps ($\approx 2.8$ epochs) with the AdamW 8\nobreakdash-bit optimizer~\citep{dettmers20228bitoptimizersblockwisequantization} ($\beta = (0.9, 0.999), \epsilon=1e^{-8}$) and a batch size of 128 on completions only. We use gradient clipping with a maximum gradient norm of $1$ and weight decay is set to $1e^{-4}$.  Furthermore, our models are fine-tuned with instruction-tuning on completions only with rank-stabilized LoRA~\citep{kalajdzievski2023rankstabilizationscalingfactor} targeting all linear layers with $\alpha_{\text{LoRA}} = 64$, $\text{dropout}_{\text{LoRA}} = 0.1$, $r_{\text{LoRA}} = 64$, and fast SVD-based PISSA~\citep{meng2024pissaprincipalsingularvalues} with 32 iterations to initialize adapter weights. 

Inspired by~\citet{hu2024minicpmunveilingpotentialsmall}, who proposed to replace linear decay with a cosine cyclic decay to increase the duration of higher learning rates, we adopt a learning rate scheduler with a stable phase of 1,000 steps with learning rate $\alpha_1 = 1e^{-4}$, a decay phase of 1,800 steps corresponding to 0.25 cosine cycles to reduce the learning rate to $\alpha_2 = 5e^{-6}$, and another smoothing decay phase of 200 steps corresponding to the remaining 0.25 cosine cycles to reduce the learning rate to $\alpha_3 = 1e^{-6}$. This increases the duration of high learning rates even further.

\section{Automated Evaluation}
\label{sec:automated-metrics-explanation}
For evaluation, we use the code provided by the \textit{Discharge Me!} challenge, which employs a comprehensive set of metrics to assess lexical and semantic similarity, factual consistency, as well as the clinical relevance and correctness.

The metrics include: \textbf{BLEU-4}~\citep{papineni-etal-2002-bleu}, which measures the precision of four-word sequences (4-grams) in the generated text against reference texts, capturing the overlap of these sequences.
\textbf{ROUGE-1, ROUGE-2, ROUGE-L}~\citep{lin-2004-rouge}, which evaluate the recall of unigrams, bigrams, and the longest common subsequence between the generated and reference texts, indicating the similarity of content.
\textbf{BERTScore}~\citep{zhang2020bertscoreevaluatingtextgeneration}, which uses contextual embeddings from BERT to evaluate the semantic similarity between the generated and reference texts.
\textbf{Meteor}~\citep{banerjee-lavie-2005-meteor}, which considers synonyms and stemming to compare the generated text with reference texts, providing a more flexible measure of similarity. 
\textbf{AlignScore}~\citep{zha-etal-2023-alignscore}, which aligns generated and reference texts to measure the quality of the alignment, reflecting the coherence and consistency of the generation.
\textbf{MEDCON}~\citep{Yim2023AcibenchAN}, which is specifically designed for medical contexts, and evaluates the clinical relevance and correctness of the generated text.

To simulate user control for automated evaluation on the \textit{DischargeMe!} dataset, we employ Llama 3.1 70B Instruct again to automatically generate authoring guidelines (Sec.~\ref{sec:authoring-guidelines}) and provide topic-level control. The LLM serves as proxy for the original authors based on the assumption that their generated output approximates the input and feedback the authors would have provided if they had originally used our methods to write the target texts $t_i$.

While authoring guidelines can be seamlessly incorporated at inference time for \textbf{w/\textsc{style}} and \textbf{w/\textsc{instr}}, simulating granular, interactive topic-level control for \textbf{w/\textsc{topics}}, that iteratively refines model output, is more complex. However, while increased user interaction generally improves output quality, it also amplifies user contribution, making it less reflective of the model's standalone performance. To minimize user contribution, we provide topic guidance indirectly and non-interactively, and effectively establish a lower-bound baseline for performance. 

Specifically, we extend the user prompt $user_i(c,g)$ with an instruction to cover a predefined list of topics (cf. Fig.~\ref{fig:evaluation_prompt_template}). This list is derived from topic segmentations (Sec.~\ref{sec:topic-level-generation-control}) for each target text $t_i$ by extracting and concatenating the headings $\mathring{h}_i^1$, ..., $\mathring{h}_i^n$ into an unnumbered bullet list.

\begin{figure}[t!]
\centering
\begin{tabular}{p{0.9\columnwidth}}
\hline
User Message \\
\hline
\vspace{-1em}
\begin{lstlisting}
{{discharge summary}}
{{radiology report 1}}
         ...
{{radiology report n}}
{{brief hospital course}}
{{authoring guidelines}}
{{instructions}}
{{topics}}
\end{lstlisting} \vspace{-1.5em} \\
\hline
\end{tabular}

\caption{The user prompt used for evaluation.}
\label{fig:evaluation_prompt_template}
\end{figure}

\section{Comparative Analysis with Existing Approaches}
\label{sec:comparative-analysis-with-existing-approaches}

\begin{table*}[ht]
\centering
\begin{tabular}{lccc}
\hline
 & BASE (ours) & WisPerMed & aehrc \\
Score & \underline{\textbf{0.363}} & \textbf{0.253} & \textbf{0.235} \\
\hline
Clinical Context & $ds + rr$ & $ds$ & optimized \\
Models trained & 1 & 2 & 2 \\
Decoding Strategy & greedy & optimized nucleus sampling & 4-beam search \\
\hline
Optimizer & AdamW 8-Bit & AdamW 8-Bit & Adam \\
Epochs & 2.8 & 3 & 5 \\
Batch Size & 128 & 16 & 16 \\
Learning Rate $\alpha$ & $1e^{-4}$ & $2e^{-4}$ & $2e^{-4}$ \\
Weight Decay & $1e^{-4}$  & $1e^{-2}$  & N/A \\
Learning Rate Scheduler & optimized WSD & linear &  linear \\
Warmup Steps & 0 & 5  & 3\% \\
\hline
LoRA Type & rank-stabilized LoRA & LoRA & QLoRA \\
Layers & all linear & all linear & all linear \\
Total Trainable Parameters & 168M & 84M & 336M \\
Weight Initialization & fast SVD-PISSA ($n=32$) & N/A & QLoRA \\
$r_{LoRA}$ & 64 & 16 & 64 \\
$\alpha_{LoRA}$ & 64 & 16 & 16 \\
dropout$_{LoRA}$ & 0.1 & 0 & N/A \\
\hline
\end{tabular}
\caption{Detailed comparison of training configurations, decoding strategies, and scores for instruction-tuned Llama 3 8B Instruct models, highlighting the key differences among our, WisPerMed's and aehrc's approach. $ds$ = Discharge Summary. $rr$ = Radiology Reports. N/A = Not Available.}
\label{tab:method-comparison}
\end{table*}

We present a detailed comparison of our instruction-tuned Llama 3 8B Instruct BASE model against the three top-performing systems on the \textit{DischargeMe!} leaderboard. We also include a detailed comparison of other instruction-tuned Llama 3 8B Instruct models evaluated during experimentation by leaderboard participants but ultimately dropped due to suboptimal performance. Table~\ref{tab:method-comparison} summarizes the primary distinctions across these methods.

\textbf{WisPerMed} \citep{damm-etal-2024-wispermed} achieved the highest leaderboard score of $0.332$, surpassing other submissions by a notable margin. This success was attributed to its Dynamic Expert Selection (DES) strategy, which combines predictions from five instruction-tuned models: Llama 3 8B and 70B Instruct, OpenBioLLM 70B, Mistral 7B Instruct (v0.2), and Phi 3 Mini 128K Instruct.  Notably, the standalone Llama 3 8B Instruct model within this ensemble achieved the lowest score ($0.253$), marginally underperforming the Phi 3 Mini model.

All models in WisPerMed's ensemble were fine-tuned using the entire discharge summary as input and LoRA with a rank of $r_{LoRA} = 16$, applied to all linear layers. In addition, some models were fine-tuned on Asclepius~\citep{kweon-etal-2024-publicly}.  For inference, they employed optimized nucleus sampling to enhance output quality. This ensemble approach enabled WisPerMed to leverage complementary model strengths, albeit at the cost of increased complexity and resource demands.

\textbf{aehrc}~\citep{liu-etal-2024-e}, similarly, fine-tuned the Llama 3 8B Instruct model using LoRA ($r_{\text{LoRA}} = 64$), but introduced notable variations in preprocessing and decoding strategies. Discharge summaries were partitioned into: (1) the text preceding the Brief Hospital Course (BHC) section for BHC generation, and (2) the text between the BHC and Discharge Instructions (DI) sections, joined with the BHC section, for DI generation. This design was motivated by their observation that longer input contexts negatively impacted model performance. They also reported that providing the entire discharge summary, including all radiology reports (as used in our setting), yielded the lowest results. For decoding, aehrc employed a 4-beam search strategy. Their leaderboard submission leveraged PRIMERA~\citep{xiao-etal-2022-primera}, a specialized instruction-tuned summarization model with 447M parameters.

\textbf{HarmonAiLab@Yale}~\citep{socrates-etal-2024-yale} has not experimented with Llama 3 8B Instruct, but GPT-3 and GPT-4 instead. 
They ultimately submitted a fine-tuned clinical model (BioBART-Large, 406M parameters) trained on an extended dataset that reportedly included samples from the validation and phase 1 test splits for a total of 83.475 ($+21.4\%$) training samples for the BHC task. HarmonAiLab@Yale also employed a 4-beam search strategy for generation, but blocking repeats with an n-gram size of 3.

\textbf{All teams} (WisPerMed, aehrc, and HarmonAiLab@Yale) trained separate models for BHC and DI tasks, effectively doubling their total trainable parameter size.

\textbf{In contrast}, we adopt a unified strategy, training a single Llama 3 8B Instruct model to jointly handle both BHC and DI tasks. The input includes the entire discharge summary and all radiology reports. Similar to aehrc, we applied LoRA ($r_{\text{LoRA}} = 64$) to all linear layers during fine-tuning, resulting in a total trainable parameter count of 168M -- double that of WisPerMed but only half that of aehrc when comparing the fine-tuned Llama 3 8B Instruct models (rather than final submissions). For decoding, we employed a greedy decoding strategy. See Table~\ref{tab:method-comparison} for a more detailed comparison.

Despite the less favorable input context and decoding strategy, our model achieved a leaderboard score of $0.363$ --- a 43\% improvement over WisPerMed's attempts, and a 54\% improvement over aehrc's attempts with instruction-tuned Llama 3 8B Instruct models (cf. Tab.~\ref{tab:method-comparison}). Moreover, our method~(168M) has significantly fewer total trainable parameters than the final submissions of WisPerMed~(1.046B), HarmonAiLab@Yale's~(812M) and aehrc's~(894M), requires less training~(2.8~epochs) than WisPerMed~(3~epochs) and aehrc~(5~epochs), and no additional data~(WisPerMed), nor an extended dataset~(HarmonAiLab@Yale). Considering all individual training setups, our training also requires only $56\%$ of Yale’s compute budget, $23\%$ of aehrc’s, and $32\%$ of WisPerMed’s

The results underscore the efficiency and effectiveness of our approach, demonstrating that instruction-tuning a single general-purpose model can achieve state-of-the-art performance without the complexity of ensembles or reliance on domain-specific models and architectures.

\section{Topic Segmentation Post-Processing}
\label{sec:topic-segmentation-post-processing}

\begin{table*}[h!]
\centering
\resizebox{\textwidth}{!}{
\begin{tabular}{lccccccccc}
\hline
  & Overall & BLEU & R-1 & R-2 & R-L & BS & Meteor & AS & MEDCON \\
\hline
Llama 3.3 70B Instruct (0-shot) & 0.175 & 0.012 & 0.286 & 0.069 & 0.142 & 0.209 & 0.205 & 0.242 & 0.239 \\
\hspace{3mm} w/\textsc{style} & 0.184 & 0.014 & 0.300 & 0.070 & 0.148 & 0.231 & 0.220 & 0.243 & 0.246 \\
\hspace{3mm} w/\textsc{instr} & 0.210 & 0.021 & 0.333 & 0.087 & 0.170 & 0.276 & 0.246 & 0.269 & 0.279 \\
\hspace{3mm} w/\textsc{topics} & 0.226 & 0.027 & 0.352 & 0.095 & 0.183 & 0.292 & 0.283 & 0.260 & 0.320 \\
\hspace{3mm} w/\textsc{style} w/\textsc{topics} & 0.225 & 0.027 & 0.348 & 0.094 & 0.182 & 0.291 & 0.281 & 0.259 & 0.319 \\
\hspace{3mm} w/\textsc{instr} w/\textsc{topics} & 0.230 & 0.027 & 0.352 & 0.097 & 0.185 & 0.296 & 0.287 & 0.270 & 0.322 \\
\hline
Llama 3.3 70B Instruct (3-shot) & 0.210 & 0.025 & 0.336 & 0.094 & 0.177 & 0.281 & 0.232 & 0.246 & 0.285 \\
\hspace{3mm} w/\textsc{style} & 0.215 & 0.027 & 0.342 & 0.097 & 0.183 & 0.296 & 0.235 & 0.254 & 0.289 \\
\hspace{3mm} w/\textsc{instr} & 0.223 & 0.029 & 0.345 & 0.104 & 0.193 & 0.317 & 0.237 & 0.259 & 0.299 \\
\hspace{3mm} w/\textsc{topics} & 0.227 & 0.027 & 0.352 & 0.095 & 0.183 & 0.292 & 0.283 & 0.261 & 0.321 \\
\hspace{3mm} w/\textsc{style} w/\textsc{topics} & 0.225 & 0.027 & 0.348 & 0.094 & 0.182 & 0.291 & 0.281 & 0.259 & 0.319 \\
\hspace{3mm} w/\textsc{instr} w/\textsc{topics} & 0.230 & 0.028 & 0.352 & 0.097 & 0.185 & 0.296 & 0.287 & 0.271 & 0.322 \\
\hline
\end{tabular}
}
\caption{The average scores per metrics for our evaluations, broken down by the two tasks: discharge instructions (DI) generation and brief hospital course (BHC) generation.}
\label{table:task_performance_llama}
\end{table*}

This step is applied only when the segmentation introduces minor alterations to the original text to avoid introducing inconsistencies between headings, questions, and text blocks through such replacements.  To achieve this, we use diff methods to identify word-level differences --- defined as whitespace-delimited character sequences --- between the generated text $\mathring{t}_i = (\mathring{t}_i^1, \dots, \mathring{t}_i^n)$ and original text $t_i$. Segmentations containing consecutive differences are then filtered out, ensuring that segmentations involving only minor differences, such as the spelling or formatting, are considering for this post-processing step. This leaves us with 93.61\% of the DI, and 81.15\% of the BHC segmentations, whose blocks $t_i^k$ are then mapped back to the original text $t_i$ to retrieve the original character sequences corresponding to each block.

\section{Human Evaluation of Topic Segmentations}
\label{sec:human-evaluation-of-topic-segmentations}

\begin{table*}[ht]
\centering
\begin{tabular}{l|ccc}
\hline
 & DS + RRs & DI & BHC  \\
\hline
\texttt{\#tokens} & 3883.76 ($\pm$ 2262.69) & 278.68  ($\pm$ 220.91) & 525.05 ($\pm$ 386.98) \\
\texttt{\#tokens(SG)} & --- & 330.62 ($\pm$ 34.51) & 325.54 ($\pm$ 29.73) \\
\texttt{\#tokens(WI)} & --- & 470.58 ($\pm$ 42.21) & 460.46 ($\pm$ 37.19) \\
\hline
\end{tabular}
\caption{Averages (and standard deviations) of token counts  for various quantities of the augmented DischargeMe! training split. Abbreviations: SG = Style Guidelines. WI = Writing Instructions. DS = Discharge Summary. RRs = Radiology Reports. DI = Discharge Instructions. BHC = Brief Hospital Course.}
\label{table:dataset_statistics}
\end{table*}

\begin{table}[ht]
\centering
\resizebox{\columnwidth}{!}{
\begin{tabular}{l|cc}
\hline
 & DI & BHC  \\
\hline
\texttt{\#segments} & 6.25  ($\pm$ 2.09) & 8.25 ( $\pm$ 4.03) \\
\texttt{\#tokens($\mathring{h}_i^k$)} & 3.82  ($\pm$ 2.00) & 4.65  ($\pm$ 2.95) \\
\texttt{\#tokens($\mathring{q}_i^k$)}  & 9.60  ($\pm$ 2.64) & 11.46 ($\pm$ 2.97) \\
\texttt{\#tokens($t_i^k$)}  & 44.57 ($\pm$ 46.16 )& 62.51 ($\pm$ 61.43) \\

\hline
\end{tabular}
}
\caption{Averages (and standard deviations) of various quantities of topic segmentations for DI and BHC sections of the DischargeMe! training split. The statistics for the number of segments \texttt{\#segments} and the token counts \texttt{\#tokens($\cdot$)} of headings $\mathring{h}_i^k$, questions $\mathring{q}_i^k$ and text blocks $t_i^k$ are consistently larger for BHC sections.}
\label{table:topic_segmentations_statistics}
\end{table}

We conducted a human evaluation of the topic segmentations (Sec.~\ref{sec:topic-level-generation-control}) generated using Llama 3.1 70B Instruct. Specifically, 500 DI and BHC sections were randomly sampled from the post-processed subset of the training split of the \textit{DischargeMe!} dataset, resulting in a total of 1000 target texts $t_i$. For each $t_i$, one segment $seg_i^j = (h_i^j, q_i^j, t_i^j)$ was randomly selection for assessment. A human expert then evaluated the selected segment through a two-step process, details in Section~\ref{sec:human-evaluation-of-topic-segmentations}.

\textbf{Step 1.} The expert was presented with the target text $t_i$, where the text range from the start of the selected text block $t_i^j$ to the end of $t_i$ was highlighted. The expert was instructed to annotate the next topic beginning within the highlighted range by identifying the heading, question, and corresponding text block. This step ensured that the expert interacted thoroughly with the target text and independently assessed and annotated the next segment without being influenced by the LLM-generated output.

\textbf{Step 2.} The expert was then provided with the LLM-generated annotation $\mathring{seg}_i^j$, which included the heading $\mathring{h}_i^j$, question $\mathring{q}_i^j$, and text block $t_i^j$. The expert evaluated the appropriateness of the generated heading, the quality of the question, and the accuracy of the text block boundaries.  While the expert could refer to their own annotations for comparison, they were instructed to assess the LLM-generated segment for correctness without imposing personal preferences, given the inherent subjectivity of the topic segmentation task and the existence of multiple competing solutions. 

The heading $\mathring{h}_i^j$ was considered appropriate only if it effectively encapsulated the content and focus of the corresponding text block $t_i^j$. The question $\mathring{q}_i^j$ was considered high quality only if it was directly answerable by the selected text block $t_i^j$ and accurately reflected and inquired the central issue or information addressed within that range.

For evaluating the text range, the expert was tasked with envisioning the optimal segment boundaries, aligning with both the heading $\mathring{h}_i^j$ and the Question Under Discussion (QUD) $\mathring{q}_i^j$, within the entire target text $t_i$. The text range was considered accurate only when the start and end points coincided with the hypothesized segment boundaries.

\textbf{Results} The evaluation revealed that the LLM-generated headings ($\mathring{h}_i^j$) aligned with the corresponding text blocks ($t_i^j$) in the majority of cases ($91.9\%$). Similarly, the generated questions ($\mathring{q}_i^j$) were well-formulated in $88.4\%$ of instances, effectively inquiring about the content of the text block and being answerable by it. In $87.5\%$ cases, both the heading and question was deemed appropriate. The accuracy of the selected text range ($t_i^j$) was confirmed in $75.2\%$ of all cases. Notably, in instances where both the headings and questions were appropriate, the accuracy of the text ranges increased to $80.91\%$.

\section{Dataset \& Annotation Statistics, Prompts and Examples}
In this section, we present examples of data augmentations, showcasing annotation samples alongside corresponding LLM prompts.
Table~\ref{table:dataset_statistics} summarizes token length statistics for the DischargeMe! training split. We find that style guidelines and writing instructions have similar average lengths across tasks (DI vs. BHC), but writing instructions are nearly 1.5× longer than style guidelines. Additionally, BHC sections are, on average, twice as long as DI sections, and BHC topic segmentations consistently contain slightly more segments, longer headings, and extended text blocks, as detailed in Table~\ref{table:topic_segmentations_statistics}.
 
\begin{table*}[ht]
\centering
\vspace{-3em}
\resizebox{0.96\textwidth}{!}{
\begin{tabular}{p{\textwidth}}
\hline
 \textbf{Synthetic Clinical Document} \\
\hline
\textcolor{MutedCoral}{The 75-year-old male patient with multi-organ sarcoidosis, diabetes mellitus, and chronic renal failure was admitted due to fatigue, dyspnea, lower limb edema, and pain.}
\textcolor{SeafoamGreen}{He received corticosteroid therapy for two years but experienced a bloodstream infection caused by Pseudomonas aeruginosa, which was successfully treated with levofloxacin. The patient's dosage of methylprednisolone was increased, leading to him being transferred to ICU and intubated due to worsening functional status.}
\textcolor{DustyBlue}{He was diagnosed with Candida albicans on Day +3 and started antifungal therapy with fluconazole (400 mg daily) and then later found to have disseminated cryptococcal disease on Day +5, leading to antifungal therapy with liposomal amphotericin B (80 mg daily).}
\textcolor{GoldenSand}{Unfortunately, the patient died from septic shock on Day +10. The laboratory findings indicated lymphocytopenia of 900 cells/µL, creatinine of 1.73 mg/dL, C-reactive protein of 83 mg/L, procalcitonin of 2.5 ng/L, increased C-reactive protein to 160 mg/L, increased procalcitonin to 14 ng/mL, and serum positive titers for CrAg ($\geq 1:4096$).}
\textcolor{BlushPink}{The diagnostic findings included pulmonary infiltration with lymphadenopathy, multiple nodules within the lung parenchyma, and disseminated cryptococcal disease.}
\textcolor{TealBreeze}{The treatment consisted of broad-spectrum antibiotic therapy with meropenem and teicoplanin, antifungal therapy with fluconazole (Day +3), and antifungal therapy with liposomal amphotericin B (Day +6).}
\textcolor{MistyLavender}{There is no follow-up information available.} 
\\
\hline
\textbf{Topic Segmentation Prompt}\\
\hline
You are tasked with fine-grained topic segmentation. Given this formatted text, segment the paragraphs into as many short blocks as sensible, each with a distinct topic. Give each block a meaningful, short topic heading, summarizing the most important information from the beginning of the block for the intended audience, and a subtitle, which reformulates the topic as a question that is answered by the block.\newline Guidelines:\newline  - Segment everything from the very first to the very last word/character/symbol.\newline - Terminate spans and insert new headings, whenever the upcoming text does not match the current running topic anymore, e.g. whenever the medical, clinical or healthcare focus changes.\newline - When formulating questions, do not use pronouns as the subjects, and do not use possessive pronouns.\newline - Do not alter the text. Copy typos, errors, mistakes and formatting from the original text.\newline  - Include headings, symbols, separators, vertical/horizontal spacing, empty lines and other formattings with their associated blocks.\newline Answer format: '<split-text>\textbackslash n\allowbreak<topic>...\allowbreak</topic>\textbackslash n\allowbreak<question>...\allowbreak</question>\textbackslash n\allowbreak<span>...\allowbreak</span>\textbackslash n\textbackslash n\allowbreak<topic>...\allowbreak</topic>\textbackslash n\allowbreak<question>...\allowbreak</question>\textbackslash n\allowbreak<span>...\allowbreak</span>\textbackslash n\textbackslash n\allowbreak...\allowbreak</split-text>' \\
\hline
 \textbf{Segmented Text} \\
\hline
\textbf{<topic>}Initial Patient Information\textbf{</topic>}

\textbf{<question>}What is the patient's initial condition and medical history?\textbf{</question>}

\textbf{<span>}\textcolor{MutedCoral}{The 75-year-old male patient with [...] fatigue, dyspnea, lower limb edema, and pain.}\textbf{</span>}

\textbf{<topic>}Initial Treatment and Complications\textbf{</topic>}

\textbf{<question>}What treatment did the patient receive initially, and what complications arose?\textbf{</question>}

\textbf{<span>}\textcolor{SeafoamGreen}{He received corticosteroid therapy for  [...] intubated due to worsening functional status.}\textbf{</span>}

\textbf{<topic>}Antifungal Therapy and Diagnosis\textbf{</topic>}

\textbf{<question>}What antifungal therapy was administered, and what diagnoses were made?\textbf{</question>}

\textbf{<span>}\textcolor{DustyBlue}{He was diagnosed with Candida albicans [...] liposomal amphotericin B (80 mg daily).}\textbf{</span>}

\textbf{<topic>}Outcome and Laboratory Findings\textbf{</topic>}

\textbf{<question>}What was the patient's outcome, and what laboratory findings were reported?\textbf{</question>}

\textbf{<span>}\textcolor{GoldenSand}{Unfortunately, the patient died from septic shock [...] positive titers for CrAg ($\geq 1:4096$).}\textbf{</span>}

\textbf{<topic>}Diagnostic Findings\textbf{</topic>}

\textbf{<question>}What diagnostic findings were reported?\textbf{</question>}

\textbf{<span>}\textcolor{BlushPink}{The diagnostic findings included [...] parenchyma, and disseminated cryptococcal disease.}\textbf{</span>}

\textbf{<topic>}Treatment Summary\textbf{</topic>}

\textbf{<question>}What treatment did the patient receive?\textbf{</question>}

\textbf{<span>}\textcolor{TealBreeze}{The treatment consisted of broad-spectrum [...] liposomal amphotericin B (Day +6).}\textbf{</span>}

\textbf{<topic>}Follow-up Information\textbf{</topic>}

\textbf{<question>}Is there any follow-up information available?\textbf{</question>}

\textbf{<span>}\textcolor{MistyLavender}{There is no follow-up information available.}\textbf{</span>}\\
\hline
\end{tabular}
}

\caption{Topic segmentation of our framework of an arbitrary clinical document retrieved from the synthetic Asclepius dataset~\cite{kweon-etal-2024-publicly} for demonstration purposes, as the \textit{DischargeMe!} dataset cannot be used directly due to privacy restrictions and access limitations. The control sequences to initiate and stop generation of specific elements are indicated in bold.}
\label{table:topic_segmentation_annotation_example}
\end{table*}

\begin{table*}[ht]
\centering
\resizebox{0.96\textwidth}{!}{
\begin{tabular}{p{\textwidth}}

\hline
\textbf{Prompt for Style Guideline Generation}\\
\hline
<text>\texttt{\{\{target text\}\}}</text>

Describe the text's tone, writing style, document format, layout, composition, textual structure, use of language, use of abbreviations, use of medical jargon, the intendened audience and anything else noteworthy. Write full sentences and paragraphs.\newline \newline Guidelines:\newline - Do not use the terms from the text.\newline  - Do not quote the text.\newline - Do not give examples from the text.\newline - Do not reveal details about the patient.\\
\hline
\textbf{Style Guideline}\\
\hline
The original text's tone is formal, objective, and neutral, indicating a professional and clinical approach to documenting a patient's medical history and treatment. The writing style is concise, direct, and informative, suggesting a focus on conveying complex medical information in a clear and efficient manner.

The document format appears to be a medical case report or a patient summary, likely used for educational, research, or clinical purposes. The layout is structured and organized, with a logical flow of information that facilitates easy comprehension. The composition is dense and detailed, indicating a high level of technical expertise and a focus on accuracy.

The textual structure is chronological, with events and treatments presented in a linear sequence. This structure allows the reader to follow the patient's progression and the corresponding medical interventions. The use of language is technical and specialized, with a high degree of medical terminology and jargon. This suggests that the intended audience is composed of medical professionals or individuals with a strong background in healthcare.

The use of abbreviations is frequent, indicating a familiarity with medical shorthand and a desire to convey complex information concisely. The text assumes a high level of prior knowledge in medicine, with minimal explanation of technical terms or concepts. This suggests that the intended audience is composed of specialists or experts in the field.

The text's use of medical jargon and technical terminology creates a sense of authority and expertise, but may also limit its accessibility to non-medical readers. The language is precise and formal, with a focus on conveying complex information in a clear and concise manner.

Overall, the text appears to be written for an audience of medical professionals, researchers, or students in the healthcare field. The tone, writing style, and use of language all suggest a focus on conveying complex medical information in a clear and efficient manner, with a high degree of technical expertise and accuracy.
\\
\hline
\end{tabular}
}

\caption{The Style Guideline generated for the synthetic clinical document from Tab.~\ref{table:topic_segmentation_annotation_example}, constituting of a list of descriptions of stylistic features.}
\label{table:style_guideline_example}
\end{table*}

\begin{table*}[ht]
\centering
\vspace{-3em}
\resizebox{0.9\textwidth}{!}{
\begin{tabular}{p{1.1\textwidth}}
\hline
\textbf{Prompt for  Writing Instructions Generation}\\
\hline
<text>\texttt{\{\{target text\}\}}</text>

Please provide detailed and comprehensive writing instructions for a non-specialist to exactly reproduce the text above. The instructions should include details on:\newline - the purpose and intent of the text, including how it is achieved\newline  - the intended audience, including how the audience's needs are met\newline  - the tone of text, including how to achieve it\newline  - the text structure and outline\newline  - the text disposition\newline  - the text formatting (not typographical), such as (but not only) the use of paragraphs, subheadings, introductions, closings, bullet points, list, including any apparent rules and patterns\newline  - the use of language, including the use of abbreviations and medical jargon (if it is used), with respect to the audience\newline  - and any other noteworty features.\newline\newline Guidelines:\newline - Use an instructive tone for writing.\newline - Consider that the non-specialist will not see the original text.\newline  - Do not use the terms from the text.\newline  - Do not quote the text.\newline  - Do not give examples from the text.\newline  - Do not reveal details about the patient.\newline\newline Answer format: '\#\# Writing Instructions\textbackslash n\textbackslash n...'
\\
\hline
 \textbf{Writing Instructions} \\
\hline
\textbf{\#\# Writing Instructions}

\textbf{\#\#\# Purpose and Intent}

The purpose of this text is to document a patient's medical history, diagnosis, treatment, and outcome in a clear and concise manner. The intent is to provide a comprehensive record of the patient's case, which can be used for medical reference, education, or research purposes. To achieve this, focus on presenting the information in a logical and chronological order, using specific details and medical terminology.

\textbf{\#\#\# Intended Audience}

The intended audience appears to be medical professionals, researchers, or students who require in-depth information about the patient's case. To meet the audience's needs, use technical language and assume a certain level of medical knowledge. However, avoid using overly complex terminology or jargon that might be unfamiliar to non-specialists in the field.

\textbf{\#\#\# Tone}

The tone of the text is formal, objective, and informative. To achieve this tone, maintain a neutral and detached perspective, focusing on presenting the facts of the case without emotional or personal commentary. Use a professional and respectful language throughout the text.

\textbf{\#\#\# Text Structure and Outline}

The text follows a chronological structure, starting with the patient's admission and progressing through their diagnosis, treatment, and outcome. The outline appears to be:

1. Introduction: Brief overview of the patient's condition and admission

2. Medical history: Description of the patient's pre-existing conditions and previous treatments

3. Diagnosis: Presentation of the patient's symptoms, laboratory findings, and diagnostic results

4. Treatment: Description of the treatments administered, including medications and therapies

5. Outcome: Report of the patient's outcome, including any complications or results

\textbf{\#\#\# Text Disposition}

The text is written in a straightforward and factual style, with no apparent attempt to persuade or entertain the reader. The focus is on conveying information in a clear and concise manner.

\textbf{\#\#\# Text Formatting}

The text appears to be written in a single block of paragraphs, with no subheadings, bullet points, or lists. The paragraphs are relatively short, with each one focusing on a specific aspect of the patient's case. To replicate this format, use a single block of text, dividing it into paragraphs as needed to separate distinct sections of information.

\textbf{\#\#\# Language and Terminology}

The text uses medical terminology and abbreviations, assuming a certain level of familiarity with medical concepts. To replicate this, use technical language and abbreviations commonly used in medical contexts. However, avoid using overly complex or specialized terminology that might be unfamiliar to non-specialists in the field.

\textbf{\#\#\# Noteworthy Features}

The text includes specific details about the patient's laboratory findings, medications, and treatments, which suggests a focus on accuracy and precision. To replicate this, include specific details and data wherever possible, using precise language and terminology.\\
\hline
\end{tabular}
}

\caption{The Writing Instructions generated for the synthetic clinical document from Tab.~\ref{table:topic_segmentation_annotation_example} alongside the respective prompt. Writing Instructions are structured and more comprehensive than Style Guidelines. They also feature a slight instructional tone.}
\label{table:writing_instructions_example}
\end{table*}

%% file: main.bbl
\begin{thebibliography}{65}
\providecommand{\natexlab}[1]{#1}

\bibitem[{Alfarghaly et~al.(2021)Alfarghaly, Khaled, Elkorany, Helal, and Fahmy}]{alfarghaly2021automated}
Omar Alfarghaly, Rana Khaled, Abeer Elkorany, Maha Helal, and Aly Fahmy. 2021.
\newblock \href {https://doi.org/10.1016/j.imu.2021.100557} {Automated radiology report generation using conditioned transformers}.
\newblock \emph{Informatics in Medicine Unlocked}, 24:100557.

\bibitem[{Ando et~al.(2022)Ando, Okumura, Komachi, Horiguchi, and Matsumoto}]{ando2022artificial}
Kenichiro Ando, Takashi Okumura, Mamoru Komachi, Hiromasa Horiguchi, and Yuji Matsumoto. 2022.
\newblock \href {https://doi.org/10.1371/journal.pdig.0000158} {Is artificial intelligence capable of generating hospital discharge summaries from inpatient records?}
\newblock \emph{PLOS Digital Health}, 1(12):e0000158.

\bibitem[{Anil et~al.(2023)Anil, Dai, Firat, Johnson, Lepikhin, Passos, Shakeri, Taropa, Bailey, Chen et~al.}]{anil2023palm2technicalreport}
Rohan Anil, Andrew~M Dai, Orhan Firat, Melvin Johnson, Dmitry Lepikhin, Alexandre Passos, Siamak Shakeri, Emanuel Taropa, Paige Bailey, Zhifeng Chen, et~al. 2023.
\newblock \href {https://doi.org/10.48550/2305.10403} {Palm 2 technical report}.
\newblock \emph{arXiv preprint arXiv:2305.10403}.

\bibitem[{Banerjee and Lavie(2005)}]{banerjee-lavie-2005-meteor}
Satanjeev Banerjee and Alon Lavie. 2005.
\newblock \href {https://aclanthology.org/W05-0909/} {{METEOR}: An automatic metric for {MT} evaluation with improved correlation with human judgments}.
\newblock In \emph{Proceedings of the {ACL} Workshop on Intrinsic and Extrinsic Evaluation Measures for Machine Translation and/or Summarization}, pages 65--72, Ann Arbor, Michigan. Association for Computational Linguistics.

\bibitem[{Brown et~al.(2020)Brown, Mann, Ryder, Subbiah, Kaplan, Dhariwal, Neelakantan, Shyam, Sastry, Askell, Agarwal, Herbert-Voss, Krueger, Henighan, Child, Ramesh, Ziegler, Wu, Winter, Hesse, Chen, Sigler, Litwin, Gray, Chess, Clark, Berner, McCandlish, Radford, Sutskever, and Amodei}]{brown2020languagemodelsfewshotlearners}
Tom~B. Brown, Benjamin Mann, Nick Ryder, Melanie Subbiah, Jared Kaplan, Prafulla Dhariwal, Arvind Neelakantan, Pranav Shyam, Girish Sastry, Amanda Askell, Sandhini Agarwal, Ariel Herbert-Voss, Gretchen Krueger, Tom Henighan, Rewon Child, Aditya Ramesh, Daniel~M. Ziegler, Jeffrey Wu, Clemens Winter, Christopher Hesse, Mark Chen, Eric Sigler, Mateusz Litwin, Scott Gray, Benjamin Chess, Jack Clark, Christopher Berner, Sam McCandlish, Alec Radford, Ilya Sutskever, and Dario Amodei. 2020.
\newblock \href {https://doi.org/10.48550/arXiv.2005.14165} {Language models are few-shot learners}.
\newblock \emph{Preprint}, arXiv:2005.14165.

\bibitem[{Clough et~al.(2024)Clough, Sparkes, Clough, Sykes, Steventon, and King}]{clough2024transforming}
Reece Alexander~James Clough, William~Anthony Sparkes, Oliver~Thomas Clough, Joshua~Thomas Sykes, Alexander~Thomas Steventon, and Kate King. 2024.
\newblock \href {https://doi.org/10.3399/BJGPO.2023.0116} {Transforming healthcare documentation: harnessing the potential of ai to generate discharge summaries}.
\newblock \emph{BJGP open}, 8(1).

\bibitem[{Dada et~al.(2024)Dada, Bauer, Contreras, Kora{\c{s}}, Seibold, Smith, and Kleesiek}]{dada2024clue}
Amin Dada, Marie Bauer, Amanda~Butler Contreras, Osman~Alperen Kora{\c{s}}, Constantin~Marc Seibold, Kaleb~E Smith, and Jens Kleesiek. 2024.
\newblock \href {https://doi.org/10.48550/2404.04067} {Does biomedical training lead to better medical performance?}
\newblock \emph{arXiv preprint arXiv:2404.04067}.

\bibitem[{Damm et~al.(2024)Damm, Pakull, Ery{\i}lmaz, Becker, Idrissi-Yaghir, Sch{\"a}fer, Schultenk{\"a}mper, and Friedrich}]{damm-etal-2024-wispermed}
Hendrik Damm, Tabea Margareta~Grace Pakull, Bahad{\i}r Ery{\i}lmaz, Helmut Becker, Ahmad Idrissi-Yaghir, Henning Sch{\"a}fer, Sergej Schultenk{\"a}mper, and Christoph~M. Friedrich. 2024.
\newblock \href {https://doi.org/10.18653/v1/2024.bionlp-1.9} {{W}is{P}er{M}ed at {\textquotedblleft}discharge me!{\textquotedblright}: Advancing text generation in healthcare with large language models, dynamic expert selection, and priming techniques on {MIMIC}-{IV}}.
\newblock In \emph{Proceedings of the 23rd Workshop on Biomedical Natural Language Processing}, pages 105--121, Bangkok, Thailand. Association for Computational Linguistics.

\bibitem[{Dettmers et~al.(2022)Dettmers, Lewis, Shleifer, and Zettlemoyer}]{dettmers20228bitoptimizersblockwisequantization}
Tim Dettmers, Mike Lewis, Sam Shleifer, and Luke Zettlemoyer. 2022.
\newblock \href {https://arxiv.org/abs/2110.02861} {8-bit optimizers via block-wise quantization}.
\newblock \emph{Preprint}, arXiv:2110.02861.

\bibitem[{Dubey et~al.(2024)Dubey, Jauhri, Pandey, Kadian, and et~al.}]{dubey2024llama3herdmodels}
Abhimanyu Dubey, Abhinav Jauhri, Abhinav Pandey, Abhishek Kadian, and Ahmad Al-Dahle et~al. 2024.
\newblock \href {https://doi.org/10.48550/2407.21783} {The llama 3 herd of models}.
\newblock \emph{Preprint}, arXiv:2407.21783.

\bibitem[{Dubinski et~al.(2024)Dubinski, Won, Trnovec, Behmanesh, Baumgarten, Dinc, Konczalla, Chan, Bernstock, Freiman et~al.}]{dubinski2024leveraging}
Daniel Dubinski, Sae-Yeon Won, Svorad Trnovec, Bedjan Behmanesh, Peter Baumgarten, Nazife Dinc, Juergen Konczalla, Alvin Chan, Joshua~D Bernstock, Thomas~M Freiman, et~al. 2024.
\newblock \href {https://doi.org/10.1007/s00701-024-05908-3} {Leveraging artificial intelligence in neurosurgery—unveiling chatgpt for neurosurgical discharge summaries and operative reports}.
\newblock \emph{Acta neurochirurgica}, 166(1):38.

\bibitem[{Edwards et~al.(2014)Edwards, Neri, Volk, Schiff, and Bates}]{edwards2014association}
Samuel~T Edwards, Pamela~M Neri, Lynn~A Volk, Gordon~D Schiff, and David~W Bates. 2014.
\newblock \href {https://doi.org/10.1136/bmjqs-2013-002194} {Association of note quality and quality of care: a cross-sectional study}.
\newblock \emph{BMJ quality \& safety}, 23(5):406--413.

\bibitem[{Ellershaw et~al.(2024)Ellershaw, Tomlinson, Burton, Frost, Hanrahan, Khan, Horsfall, Little, Malgapo, Starup-Hansen et~al.}]{ellershaw2024automated}
Simon Ellershaw, Christopher Tomlinson, Oliver~E Burton, Thomas Frost, John~Gerrard Hanrahan, Danyal~Zaman Khan, Hugo~Layard Horsfall, Mollie Little, Evaleen Malgapo, Joachim Starup-Hansen, et~al. 2024.
\newblock \href {https://openreview.net/forum?id=1kDJJPppRG} {Automated generation of hospital discharge summaries using clinical guidelines and large language models}.
\newblock In \emph{AAAI 2024 Spring Symposium on Clinical Foundation Models}.

\bibitem[{Fakhoury et~al.(2024)Fakhoury, Naik, Sakkas, Chakraborty, and Lahiri}]{10606356}
Sarah Fakhoury, Aaditya Naik, Georgios Sakkas, Saikat Chakraborty, and Shuvendu~K. Lahiri. 2024.
\newblock \href {https://doi.org/10.1109/TSE.2024.3428972} {Llm-based test-driven interactive code generation: User study and empirical evaluation}.
\newblock \emph{IEEE Transactions on Software Engineering}, 50(9):2254--2268.

\bibitem[{Gilardi et~al.(2023)Gilardi, Alizadeh, and Kubli}]{doi:10.1073/pnas.2305016120}
Fabrizio Gilardi, Meysam Alizadeh, and Maël Kubli. 2023.
\newblock \href {https://doi.org/10.1073/pnas.2305016120} {Chatgpt outperforms crowd workers for text-annotation tasks}.
\newblock \emph{Proceedings of the National Academy of Sciences}, 120(30):e2305016120.

\bibitem[{Hartman and Campion(2022)}]{hartman2022day}
Vince Hartman and Thomas~R Campion. 2022.
\newblock \href {https://www.ncbi.nlm.nih.gov/pmc/articles/PMC9285173/} {A day-to-day approach for automating the hospital course section of the discharge summary}.
\newblock \emph{AMIA Summits on Translational Science Proceedings}, 2022:216.

\bibitem[{Hartman et~al.(2023)Hartman, Bapat, Weiner, Navi, Sholle, and Campion~Jr}]{hartman2023method}
Vince~C Hartman, Sanika~S Bapat, Mark~G Weiner, Babak~B Navi, Evan~T Sholle, and Thomas~R Campion~Jr. 2023.
\newblock \href {https://doi.org/10.1093/jamia/ocad177} {A method to automate the discharge summary hospital course for neurology patients}.
\newblock \emph{Journal of the American Medical Informatics Association}, 30(12):1995--2003.

\bibitem[{Hirosawa et~al.(2023)Hirosawa, Harada, Yokose, Sakamoto, Kawamura, and Shimizu}]{ijerph20043378}
Takanobu Hirosawa, Yukinori Harada, Masashi Yokose, Tetsu Sakamoto, Ren Kawamura, and Taro Shimizu. 2023.
\newblock \href {https://doi.org/10.3390/ijerph20043378} {Diagnostic accuracy of differential-diagnosis lists generated by generative pretrained transformer 3 chatbot for clinical vignettes with common chief complaints: A pilot study}.
\newblock \emph{International Journal of Environmental Research and Public Health}, 20(4).

\bibitem[{Hu et~al.(2024)Hu, Tu, Han, He, Cui, Long, Zheng, Fang, Huang, Zhao, Zhang, Thai, Zhang, Wang, Yao, Zhao, Zhou, Cai, Zhai, Ding, Jia, Zeng, Li, Liu, and Sun}]{hu2024minicpmunveilingpotentialsmall}
Shengding Hu, Yuge Tu, Xu~Han, Chaoqun He, Ganqu Cui, Xiang Long, Zhi Zheng, Yewei Fang, Yuxiang Huang, Weilin Zhao, Xinrong Zhang, Zheng~Leng Thai, Kaihuo Zhang, Chongyi Wang, Yuan Yao, Chenyang Zhao, Jie Zhou, Jie Cai, Zhongwu Zhai, Ning Ding, Chao Jia, Guoyang Zeng, Dahai Li, Zhiyuan Liu, and Maosong Sun. 2024.
\newblock \href {https://doi.org/10.48550/arXiv.2404.06395} {Minicpm: Unveiling the potential of small language models with scalable training strategies}.
\newblock \emph{Preprint}, arXiv:2404.06395.

\bibitem[{Huang et~al.(2024)Huang, Safranek, Socrates, Chartash, Wright, Dilip, Sangal, and Taylor}]{Huang2024PatientRepresentingPP}
Thomas Huang, Conrad~W Safranek, Vimig Socrates, David Chartash, Donald Wright, Monisha Dilip, Rohit~B. Sangal, and Richard~Andrew Taylor. 2024.
\newblock \href {https://doi.org/10.2196/60336} {Patient-representing population's perceptions of gpt-generated versus standard emergency department discharge instructions: Randomized blind survey assessment}.
\newblock \emph{Journal of Medical Internet Research}, 26.

\bibitem[{Hultman et~al.(2019)Hultman, Marquard, Lindemann, Arsoniadis, Pakhomov, and Melton}]{Hultman2019ChallengesAO}
Gretchen~M Hultman, Jenna~L Marquard, Elizabeth Lindemann, Elliot Arsoniadis, Serguei Pakhomov, and Genevieve~B Melton. 2019.
\newblock \href {https://doi.org/10.1055/s-0039-1692164} {Challenges and opportunities to improve the clinician experience reviewing electronic progress notes}.
\newblock \emph{Applied clinical informatics}, 10(03):446--453.

\bibitem[{Huot et~al.(2023)Huot, Maynez, Narayan, Amplayo, Ganchev, Louis, Sandholm, Das, and Lapata}]{huot-etal-2023-text}
Fantine Huot, Joshua Maynez, Shashi Narayan, Reinald~Kim Amplayo, Kuzman Ganchev, Annie~Priyadarshini Louis, Anders Sandholm, Dipanjan Das, and Mirella Lapata. 2023.
\newblock \href {https://doi.org/10.18653/v1/2023.eacl-demo.13} {Text-blueprint: An interactive platform for plan-based conditional generation}.
\newblock In \emph{Proceedings of the 17th Conference of the European Chapter of the Association for Computational Linguistics: System Demonstrations}, pages 105--116, Dubrovnik, Croatia. Association for Computational Linguistics.

\bibitem[{Ji et~al.(2023)Ji, Lee, Frieske, Yu, Su, Xu, Ishii, Bang, Madotto, and Fung}]{10.1145/3571730}
Ziwei Ji, Nayeon Lee, Rita Frieske, Tiezheng Yu, Dan Su, Yan Xu, Etsuko Ishii, Ye~Jin Bang, Andrea Madotto, and Pascale Fung. 2023.
\newblock \href {https://doi.org/10.1145/3571730} {Survey of hallucination in natural language generation}.
\newblock \emph{ACM Comput. Surv.}, 55(12).

\bibitem[{Kalajdzievski(2023)}]{kalajdzievski2023rankstabilizationscalingfactor}
Damjan Kalajdzievski. 2023.
\newblock \href {https://doi.org/10.48550/arXiv.2312.03732} {A rank stabilization scaling factor for fine-tuning with lora}.
\newblock \emph{Preprint}, arXiv:2312.03732.

\bibitem[{Keskar et~al.(2019)Keskar, McCann, Varshney, Xiong, and Socher}]{keskar2019ctrl}
Nitish~Shirish Keskar, Bryan McCann, Lav~R Varshney, Caiming Xiong, and Richard Socher. 2019.
\newblock \href {https://doi.org/10.48550/arXiv.1909.05858} {Ctrl: A conditional transformer language model for controllable generation}.
\newblock \emph{arXiv preprint arXiv:1909.05858}.

\bibitem[{Kweon et~al.(2024)Kweon, Kim, Kim, Im, Cho, Bae, Oh, Lee, Moon, You, Baek, Han, Jung, Jo, and Choi}]{kweon-etal-2024-publicly}
Sunjun Kweon, Junu Kim, Jiyoun Kim, Sujeong Im, Eunbyeol Cho, Seongsu Bae, Jungwoo Oh, Gyubok Lee, Jong~Hak Moon, Seng~Chan You, Seungjin Baek, Chang~Hoon Han, Yoon~Bin Jung, Yohan Jo, and Edward Choi. 2024.
\newblock \href {https://doi.org/10.18653/v1/2024.findings-acl.305} {Publicly shareable clinical large language model built on synthetic clinical notes}.
\newblock In \emph{Findings of the Association for Computational Linguistics: ACL 2024}, pages 5148--5168, Bangkok, Thailand. Association for Computational Linguistics.

\bibitem[{Lin(2004)}]{lin-2004-rouge}
Chin-Yew Lin. 2004.
\newblock \href {https://aclanthology.org/W04-1013/} {{ROUGE}: A package for automatic evaluation of summaries}.
\newblock In \emph{Text Summarization Branches Out}, pages 74--81, Barcelona, Spain. Association for Computational Linguistics.

\bibitem[{Lin et~al.(2024)Lin, Guan, Zhang, Zhang, Li, and Zhang}]{Lin2024TowardsTL}
Zichao Lin, Shuyan Guan, Wending Zhang, Huiyan Zhang, Yugang Li, and Huaping Zhang. 2024.
\newblock \href {https://doi.org/10.1007/s10462-024-10896-y} {Towards trustworthy llms: a review on debiasing and dehallucinating in large language models}.
\newblock \emph{Artif. Intell. Rev.}, 57:243.

\bibitem[{Liu et~al.(2023)Liu, Wang, and Liu}]{Liu2023UtilityOC}
Jialin Liu, Changyu Wang, and Siru Liu. 2023.
\newblock \href {https://doi.org/10.2196/48568} {Utility of chatgpt in clinical practice}.
\newblock \emph{Journal of Medical Internet Research}, 25:e48568.

\bibitem[{Liu et~al.(2024)Liu, Nicolson, Dowling, Koopman, and Nguyen}]{liu-etal-2024-e}
Jinghui Liu, Aaron Nicolson, Jason Dowling, Bevan Koopman, and Anthony Nguyen. 2024.
\newblock \href {https://doi.org/10.18653/v1/2024.bionlp-1.59} {e-health {CSIRO} at {\textquotedblleft}discharge me!{\textquotedblright} 2024: Generating discharge summary sections with fine-tuned language models}.
\newblock In \emph{Proceedings of the 23rd Workshop on Biomedical Natural Language Processing}, pages 675--684, Bangkok, Thailand. Association for Computational Linguistics.

\bibitem[{Meng et~al.(2024)Meng, Wang, and Zhang}]{meng2024pissaprincipalsingularvalues}
Fanxu Meng, Zhaohui Wang, and Muhan Zhang. 2024.
\newblock \href {https://doi.org/10.48550/arXiv.2404.02948} {Pissa: Principal singular values and singular vectors adaptation of large language models}.
\newblock \emph{Preprint}, arXiv:2404.02948.

\bibitem[{Mesk{\'o} and Topol(2023)}]{Mesk2023TheIF}
Bertalan Mesk{\'o} and Eric~J. Topol. 2023.
\newblock \href {https://api.semanticscholar.org/CorpusID:259357970} {The imperative for regulatory oversight of large language models (or generative ai) in healthcare}.
\newblock \emph{NPJ Digital Medicine}, 6.

\bibitem[{Mu et~al.(2024)Mu, Shi, Wang, Yu, Zhang, Wang, Liu, and Wang}]{10.1145/3660810}
Fangwen Mu, Lin Shi, Song Wang, Zhuohao Yu, Binquan Zhang, ChenXue Wang, Shichao Liu, and Qing Wang. 2024.
\newblock \href {https://doi.org/10.1145/3660810} {Clarifygpt: A framework for enhancing llm-based code generation via requirements clarification}.
\newblock \emph{Proc. ACM Softw. Eng.}, 1(FSE).

\bibitem[{Narayan et~al.(2023)Narayan, Maynez, Amplayo, Ganchev, Louis, Huot, Sandholm, Das, and Lapata}]{narayan-etal-2023-conditional}
Shashi Narayan, Joshua Maynez, Reinald~Kim Amplayo, Kuzman Ganchev, Annie Louis, Fantine Huot, Anders Sandholm, Dipanjan Das, and Mirella Lapata. 2023.
\newblock \href {https://doi.org/10.1162/tacl_a_00583} {Conditional generation with a question-answering blueprint}.
\newblock \emph{Transactions of the Association for Computational Linguistics}, 11:974--996.

\bibitem[{Omiye et~al.(2024)Omiye, Gui, Rezaei, Zou, and Daneshjou}]{PMID:38285984}
Jesutofunmi~A Omiye, Haiwen Gui, Shawheen~J Rezaei, James Zou, and Roxana Daneshjou. 2024.
\newblock \href {https://doi.org/10.7326/m23-2772} {Large language models in medicine: The potentials and pitfalls : A narrative review}.
\newblock \emph{Annals of internal medicine}, 177(2):210—220.

\bibitem[{OpenAI et~al.(2024)OpenAI, Achiam, Adler, Agarwal, Ahmad, and et~al.}]{openai2024gpt4technicalreport}
OpenAI, Josh Achiam, Steven Adler, Sandhini Agarwal, Lama Ahmad, and Ilge~Akkaya et~al. 2024.
\newblock \href {https://doi.org/10.48550/2303.08774} {Gpt-4 technical report}.
\newblock \emph{Preprint}, arXiv:2303.08774.

\bibitem[{Papineni et~al.(2002)Papineni, Roukos, Ward, and Zhu}]{papineni-etal-2002-bleu}
Kishore Papineni, Salim Roukos, Todd Ward, and Wei-Jing Zhu. 2002.
\newblock \href {https://doi.org/10.3115/1073083.1073135} {{B}leu: a method for automatic evaluation of machine translation}.
\newblock In \emph{Proceedings of the 40th Annual Meeting of the Association for Computational Linguistics}, pages 311--318, Philadelphia, Pennsylvania, USA. Association for Computational Linguistics.

\bibitem[{Patel and Lam(2023)}]{Patel2023ChatGPTTF}
Sajan~B Patel and Kyle Lam. 2023.
\newblock \href {https://doi.org/10.1016/S2589-7500(23)00021-3} {Chatgpt: the future of discharge summaries?}
\newblock \emph{The Lancet Digital Health}, 5(3):e107--e108.

\bibitem[{Perez et~al.(2022)Perez, Huang, Song, Cai, Ring, Aslanides, Glaese, McAleese, and Irving}]{perez-etal-2022-red}
Ethan Perez, Saffron Huang, Francis Song, Trevor Cai, Roman Ring, John Aslanides, Amelia Glaese, Nat McAleese, and Geoffrey Irving. 2022.
\newblock \href {https://doi.org/10.18653/v1/2022.emnlp-main.225} {Red teaming language models with language models}.
\newblock In \emph{Proceedings of the 2022 Conference on Empirical Methods in Natural Language Processing}, pages 3419--3448, Abu Dhabi, United Arab Emirates. Association for Computational Linguistics.

\bibitem[{Pollard et~al.(2013)Pollard, Neri, Wilcox, Volk, Williams, Schiff, Ramelson, and Bates}]{POLLARD201339}
Stephanie~E. Pollard, Pamela~M. Neri, Allison~R. Wilcox, Lynn~A. Volk, Deborah~H. Williams, Gordon~D. Schiff, Harley~Z. Ramelson, and David~W. Bates. 2013.
\newblock \href {https://doi.org/10.1016/j.ijmedinf.2012.04.002} {How physicians document outpatient visit notes in an electronic health record}.
\newblock \emph{International Journal of Medical Informatics}, 82(1):39--46.

\bibitem[{Roberts(2012)}]{roberts:2012:information}
Craige Roberts. 2012.
\newblock \href {https://doi.org/10.3765/sp.5.6} {Information structure in discourse: Towards an integrated formal theory of pragmatics}.
\newblock \emph{Semantics and Pragmatics}, 5(6):1--69.

\bibitem[{Ruinelli et~al.(2024)Ruinelli, Colombo, Rochat, Popeskou, Franchini, Mitrovi{\'c}, Lithgow, Cornelius, and Rinaldi}]{ruinelli-etal-2024-experiments}
Lorenzo Ruinelli, Amos Colombo, Mathilde Rochat, Sotirios~Georgios Popeskou, Andrea Franchini, Sandra Mitrovi{\'c}, Oscar~William Lithgow, Joseph Cornelius, and Fabio Rinaldi. 2024.
\newblock \href {https://aclanthology.org/2024.cl4health-1.17/} {Experiments in automated generation of discharge summaries in {I}talian}.
\newblock In \emph{Proceedings of the First Workshop on Patient-Oriented Language Processing (CL4Health) @ LREC-COLING 2024}, pages 137--144, Torino, Italia. ELRA and ICCL.

\bibitem[{Searle et~al.(2023)Searle, Ibrahim, Teo, and Dobson}]{searle2023discharge}
Thomas Searle, Zina Ibrahim, James Teo, and Richard~JB Dobson. 2023.
\newblock \href {https://doi.org/10.1016/j.jbi.2023.104358} {Discharge summary hospital course summarisation of in patient electronic health record text with clinical concept guided deep pre-trained transformer models}.
\newblock \emph{Journal of Biomedical Informatics}, 141:104358.

\bibitem[{Singhal et~al.(2023)Singhal, Tu, Gottweis, Sayres, Wulczyn, Hou, Clark, Pfohl, Cole-Lewis, Neal, Schaekermann, Wang, Amin, Lachgar, Mansfield, Prakash, Green, Dominowska, y~Arcas, Tomasev, Liu, Wong, Semturs, Mahdavi, Barral, Webster, Corrado, Matias, Azizi, Karthikesalingam, and Natarajan}]{singhal2023expertlevelmedicalquestionanswering}
Karan Singhal, Tao Tu, Juraj Gottweis, Rory Sayres, Ellery Wulczyn, Le~Hou, Kevin Clark, Stephen Pfohl, Heather Cole-Lewis, Darlene Neal, Mike Schaekermann, Amy Wang, Mohamed Amin, Sami Lachgar, Philip Mansfield, Sushant Prakash, Bradley Green, Ewa Dominowska, Blaise~Aguera y~Arcas, Nenad Tomasev, Yun Liu, Renee Wong, Christopher Semturs, S.~Sara Mahdavi, Joelle Barral, Dale Webster, Greg~S. Corrado, Yossi Matias, Shekoofeh Azizi, Alan Karthikesalingam, and Vivek Natarajan. 2023.
\newblock \href {https://doi.org/10.48550/arXiv.2305.09617} {Towards expert-level medical question answering with large language models}.
\newblock \emph{Preprint}, arXiv:2305.09617.

\bibitem[{Socrates et~al.(2024)Socrates, Huang, Ai, Fereydooni, Chen, Taylor, and Chartash}]{socrates-etal-2024-yale}
Vimig Socrates, Thomas Huang, Xuguang Ai, Soraya Fereydooni, Qingyu Chen, R~Andrew Taylor, and David Chartash. 2024.
\newblock \href {https://doi.org/10.18653/v1/2024.bionlp-1.64} {{Y}ale at {\textquotedblleft}discharge me!{\textquotedblright}: Evaluating constrained generation of discharge summaries with unstructured and structured information}.
\newblock In \emph{Proceedings of the 23rd Workshop on Biomedical Natural Language Processing}, pages 724--730, Bangkok, Thailand. Association for Computational Linguistics.

\bibitem[{Soleimani et~al.(2024)Soleimani, Seyyedi, Ayyoubzadeh, Kalhori, and Keshavarz}]{soleimani2024practical}
Mohsen Soleimani, Navisa Seyyedi, Seyed~Mohammad Ayyoubzadeh, Sharareh Rostam~Niakan Kalhori, and Hamidreza Keshavarz. 2024.
\newblock \href {https://doi.org/10.1016/j.acra.2024.07.020} {Practical evaluation of chatgpt performance for radiology report generation}.
\newblock \emph{Academic Radiology}.

\bibitem[{Tang et~al.(2023)Tang, Sun, Idnay, Nestor, Soroush, Elias, Xu, Ding, Durrett, Rousseau et~al.}]{tang2023evaluating}
Liyan Tang, Zhaoyi Sun, Betina Idnay, Jordan~G Nestor, Ali Soroush, Pierre~A Elias, Ziyang Xu, Ying Ding, Greg Durrett, Justin~F Rousseau, et~al. 2023.
\newblock \href {https://doi.org/1038/s41746-023-00896-7} {Evaluating large language models on medical evidence summarization}.
\newblock \emph{NPJ digital medicine}, 6(1):158.

\bibitem[{Tao et~al.(2024)Tao, Xi, Li, Tang, and Xu}]{tao2024cat}
Zhen Tao, Dinghao Xi, Zhiyu Li, Liumin Tang, and Wei Xu. 2024.
\newblock \href {https://doi.org/10.48550/arXiv.2401.05707} {Cat-llm: Prompting large language models with text style definition for chinese article-style transfer}.
\newblock \emph{arXiv preprint arXiv:2401.05707}.

\bibitem[{Team et~al.(2024)Team, Anil, Borgeaud, Alayrac, Yu, and et~al.}]{geminiteam2024geminifamilyhighlycapable}
Gemini Team, Rohan Anil, Sebastian Borgeaud, Jean-Baptiste Alayrac, Jiahui Yu, and Radu~Soricut et~al. 2024.
\newblock \href {https://doi.org/10.48550/2312.11805} {Gemini: A family of highly capable multimodal models}.
\newblock \emph{Preprint}, arXiv:2312.11805.

\bibitem[{Touvron et~al.(2023{\natexlab{a}})Touvron, Lavril, Izacard, Martinet, Lachaux, Lacroix, Rozière, Goyal, Hambro, Azhar, Rodriguez, Joulin, Grave, and Lample}]{touvron2023llamaopenefficientfoundation}
Hugo Touvron, Thibaut Lavril, Gautier Izacard, Xavier Martinet, Marie-Anne Lachaux, Timothée Lacroix, Baptiste Rozière, Naman Goyal, Eric Hambro, Faisal Azhar, Aurelien Rodriguez, Armand Joulin, Edouard Grave, and Guillaume Lample. 2023{\natexlab{a}}.
\newblock \href {https://doi.org/10.48550/arXiv.2302.13971} {Llama: Open and efficient foundation language models}.
\newblock \emph{Preprint}, arXiv:2302.13971.

\bibitem[{Touvron et~al.(2023{\natexlab{b}})Touvron, Martin, Stone, Albert, and et~al.}]{touvron2023llama2openfoundation}
Hugo Touvron, Louis Martin, Kevin Stone, Peter Albert, and Amjad~Almahairi et~al. 2023{\natexlab{b}}.
\newblock \href {https://doi.org/10.48550/arXiv.2307.09288} {Llama 2: Open foundation and fine-tuned chat models}.
\newblock \emph{Preprint}, arXiv:2307.09288.

\bibitem[{Van~Kuppevelt(1995)}]{van1995discourse}
Jan Van~Kuppevelt. 1995.
\newblock \href {https://doi.org/10.1017/S002222670000058X} {Discourse structure, topicality and questioning}.
\newblock \emph{Journal of linguistics}, 31(1):109--147.

\bibitem[{Van~Veen et~al.(2024)Van~Veen, Van~Uden, Blankemeier, Delbrouck, Aali, Bluethgen, Pareek, Polacin, Reis, Seehofnerov{\'a} et~al.}]{van2024adapted}
Dave Van~Veen, Cara Van~Uden, Louis Blankemeier, Jean-Benoit Delbrouck, Asad Aali, Christian Bluethgen, Anuj Pareek, Malgorzata Polacin, Eduardo~Pontes Reis, Anna Seehofnerov{\'a}, et~al. 2024.
\newblock \href {https://doi.org/10.1038/s41591-024-02855-5} {Adapted large language models can outperform medical experts in clinical text summarization}.
\newblock \emph{Nature medicine}, 30(4):1134--1142.

\bibitem[{wai Yim et~al.(2023)wai Yim, Fu, Abacha, Snider, Lin, and Yetisgen}]{Yim2023AcibenchAN}
Wen wai Yim, Yujuan Fu, Asma~Ben Abacha, Neal Snider, Thomas Lin, and Meliha Yetisgen. 2023.
\newblock \href {https://doi.org/10.1038/s41597-023-02487-3} {Aci-bench: a novel ambient clinical intelligence dataset for benchmarking automatic visit note generation}.
\newblock \emph{Scientific Data}, 10.

\bibitem[{Wang et~al.(2023)Wang, Liu, Wang, and Zhou}]{wang2023r2gengpt}
Zhanyu Wang, Lingqiao Liu, Lei Wang, and Luping Zhou. 2023.
\newblock \href {https://doi.org/10.1016/j.metrad.2023.100033} {R2gengpt: Radiology report generation with frozen llms}.
\newblock \emph{Meta-Radiology}, 1(3):100033.

\bibitem[{Were et~al.(2010)Were, Shen, Bwana, Emenyonu, Musinguzi, Nkuyahaga, Kembabazi, and Tierney}]{Were2010CreationAE}
Martin~Chieng Were, Changyu Shen, Mwebesa~B. Bwana, Nneka Emenyonu, Nicholas Musinguzi, Frank Nkuyahaga, Annet Kembabazi, and William~M. Tierney. 2010.
\newblock \href {https://doi.org/10.1016/j.ijmedinf.2009.11.006} {Creation and evaluation of emr-based paper clinical summaries to support hiv-care in uganda, africa}.
\newblock \emph{International journal of medical informatics}, 79 2:90--6.

\bibitem[{Xiao et~al.(2022)Xiao, Beltagy, Carenini, and Cohan}]{xiao-etal-2022-primera}
Wen Xiao, Iz~Beltagy, Giuseppe Carenini, and Arman Cohan. 2022.
\newblock \href {https://doi.org/10.18653/v1/2022.acl-long.360} {{PRIMERA}: Pyramid-based masked sentence pre-training for multi-document summarization}.
\newblock In \emph{Proceedings of the 60th Annual Meeting of the Association for Computational Linguistics (Volume 1: Long Papers)}, pages 5245--5263, Dublin, Ireland. Association for Computational Linguistics.

\bibitem[{Xu et~al.(2024)Xu, Chen, Johnston, Blankemeier, Varma, Hom, Collins, Modi, Lloyd, Hopkins, Langlotz, and Delbrouck}]{xu-etal-2024-overview}
Justin Xu, Zhihong Chen, Andrew Johnston, Louis Blankemeier, Maya Varma, Jason Hom, William~J. Collins, Ankit Modi, Robert Lloyd, Benjamin Hopkins, Curtis Langlotz, and Jean-Benoit Delbrouck. 2024.
\newblock \href {https://doi.org/10.18653/v1/2024.bionlp-1.7} {Overview of the first shared task on clinical text generation: {RRG}24 and {\textquotedblleft}discharge me!{\textquotedblright}}.
\newblock In \emph{Proceedings of the 23rd Workshop on Biomedical Natural Language Processing}, pages 85--98, Bangkok, Thailand. Association for Computational Linguistics.

\bibitem[{Yang and Klein(2021)}]{yang-klein-2021-fudge}
Kevin Yang and Dan Klein. 2021.
\newblock \href {https://doi.org/10.18653/v1/2021.naacl-main.276} {{FUDGE}: Controlled text generation with future discriminators}.
\newblock In \emph{Proceedings of the 2021 Conference of the North American Chapter of the Association for Computational Linguistics: Human Language Technologies}, pages 3511--3535, Online. Association for Computational Linguistics.

\bibitem[{Yang et~al.(2023)Yang, Wu, Ge, Zheng, Zhou, and Xiao}]{yang2023radiology}
Shuxin Yang, Xian Wu, Shen Ge, Zhuozhao Zheng, S~Kevin Zhou, and Li~Xiao. 2023.
\newblock \href {https://doi.org/10.1016/j.media.2023.102798} {Radiology report generation with a learned knowledge base and multi-modal alignment}.
\newblock \emph{Medical Image Analysis}, 86:102798.

\bibitem[{Zaretsky et~al.(2024)Zaretsky, Kim, Baskharoun, Zhao, Austrian, Aphinyanaphongs, Gupta, Blecker, and Feldman}]{Zaretsky2024GenerativeAI}
Jonah Zaretsky, Jeong~Min Kim, Samuel Baskharoun, Yunan Zhao, Jonathan~S Austrian, Yindalon Aphinyanaphongs, Ravi Gupta, Saul~B. Blecker, and Jonah Feldman. 2024.
\newblock \href {https://doi.org/10.1001/jamanetworkopen.2024.0357} {Generative artificial intelligence to transform inpatient discharge summaries to patient-friendly language and format}.
\newblock \emph{JAMA Network Open}, 7.

\bibitem[{Zha et~al.(2023)Zha, Yang, Li, and Hu}]{zha-etal-2023-alignscore}
Yuheng Zha, Yichi Yang, Ruichen Li, and Zhiting Hu. 2023.
\newblock \href {https://doi.org/10.18653/v1/2023.acl-long.634} {{A}lign{S}core: Evaluating factual consistency with a unified alignment function}.
\newblock In \emph{Proceedings of the 61st Annual Meeting of the Association for Computational Linguistics (Volume 1: Long Papers)}, pages 11328--11348, Toronto, Canada. Association for Computational Linguistics.

\bibitem[{Zhang et~al.(2023)Zhang, Song, Li, Zhou, and Song}]{zhang2023survey}
Hanqing Zhang, Haolin Song, Shaoyu Li, Ming Zhou, and Dawei Song. 2023.
\newblock \href {https://doi.org/10.1145/3617680} {A survey of controllable text generation using transformer-based pre-trained language models}.
\newblock \emph{ACM Computing Surveys}, 56(3):1--37.

\bibitem[{Zhang et~al.(2020)Zhang, Kishore, Wu, Weinberger, and Artzi}]{zhang2020bertscoreevaluatingtextgeneration}
Tianyi Zhang, Varsha Kishore, Felix Wu, Kilian~Q. Weinberger, and Yoav Artzi. 2020.
\newblock \href {https://doi.org/10.48550/arXiv.1904.09675} {Bertscore: Evaluating text generation with bert}.
\newblock \emph{Preprint}, arXiv:1904.09675.

\bibitem[{Zou et~al.(2021)Zou, Yin, Zhong, Yang, Yang, and Tang}]{zou2021controllable}
Xu~Zou, Da~Yin, Qingyang Zhong, Hongxia Yang, Zhilin Yang, and Jie Tang. 2021.
\newblock \href {https://doi.org/10.1145/3447548.3467418} {Controllable generation from pre-trained language models via inverse prompting}.
\newblock In \emph{Proceedings of the 27th ACM SIGKDD Conference on Knowledge Discovery \& Data Mining}, pages 2450--2460.

\end{thebibliography}
